%% file: sample-manuscript.tex
\documentclass[manuscript,screen]{acmart}
\usepackage{multirow}

\usepackage{amsthm}
\usepackage{graphicx}
\usepackage{graphics}
\usepackage{subfigure}
\usepackage{amsmath}
\usepackage{float}
\usepackage{algorithm}     
\usepackage{algorithmic}
\usepackage{bbding}

\AtBeginDocument{%
  \providecommand\BibTeX{{%
    \normalfont B\kern-0.5em{\scshape i\kern-0.25em b}\kern-0.8em\TeX}}}

\setcopyright{acmcopyright}
\copyrightyear{2018}
\acmYear{2018}
\acmDOI{10.1145/1122445.1122456}

\acmConference[Woodstock '18]{Woodstock '18: ACM Symposium on Neural
  Gaze Detection}{June 03--05, 2018}{Woodstock, NY}
\acmBooktitle{Woodstock '18: ACM Symposium on Neural Gaze Detection,
  June 03--05, 2018, Woodstock, NY}
\acmPrice{15.00}
\acmISBN{978-1-4503-XXXX-X/18/06}

\begin{document}

\title{DyGCN: Dynamic Graph Embedding with Graph Convolutional Network}

\author{Zeyu Cui}
\authornote{Both authors contributed equally to this research.}
\email{zeyu.cui@nlpr.ia.ac.cn}
\orcid{0000-0003-0017-5292}
\affiliation{%
  \institution{University of Chinese Academy of Sciences (UCAS)}
  \institution{Institution of Automation Chinese Academy of Sciences (CASIA)}
  \institution{National Laboratory of Pattern Recognition (NLPR)}
  \institution{Center for Research on Intelligent Perception and Computing(CRIPAC)}
  \streetaddress{No.95 ZhongGuanCun East Road}
  \postcode{100080}
  \city{Beijing}
  \country{China}
}
\author{Zekun Li}
\authornotemark[1]
\email{lizekunlee@gmail.com}
\affiliation{%
  \institution{School of Cyber Security, University of Chinese Academy of Sciences}
  \city{Beijing}
  \country{China}
  \postcode{100080}
}

\author{Shu Wu}
\email{shu.wu@nlpr.ia.ac.cn}
\affiliation{%
  \institution{University of Chinese Academy of Sciences (UCAS)}
  \institution{Institution of Automation Chinese Academy of Sciences (CASIA)}
  \institution{National Laboratory of Pattern Recognition (NLPR)}
  \institution{Center for Research on Intelligent Perception and Computing(CRIPAC)}
  \streetaddress{No.95 ZhongGuanCun East Road}
  \postcode{100080}
  \city{Beijing}
  \country{China}
}

\author{Xiaoyu Zhang}
\email{zhangxiaoyu@iie.ac.cn}
\affiliation{%
  \institution{School of Cyber Security, University of Chinese Academy of Sciences}
  \city{Beijing}
  \country{China}
  \postcode{100080}
}

\author{Qiang Liu}
\email{qiang.liu@nlpr.ia.ac.cn}

\author{Liang Wang}
\email{wangliang@nlpr.ia.ac.cn}
\affiliation{%
  \institution{University of Chinese Academy of Sciences (UCAS)}
  \institution{Institution of Automation Chinese Academy of Sciences (CASIA)}
  \institution{National Laboratory of Pattern Recognition (NLPR)}
  \institution{Center for Research on Intelligent Perception and Computing(CRIPAC)}
  \streetaddress{No.95 ZhongGuanCun East Road}
  \postcode{100080}
  \city{Beijing}
  \country{China}
}

\author{Mengmeng Ai}
\affiliation{%
  \institution{the Beijing University of Post and Telecommunication}
  \postcode{100080}
  \city{Beijing}
  \country{China}
}


\begin{abstract}
Graph embedding, aiming to learn low-dimensional representations (aka. embeddings) of nodes, has received significant attention recently. Recent years have witnessed a surge of efforts made on static graphs, among which Graph Convolutional Network (GCN) has emerged as an effective class of models. However, these methods mainly focus on the static graph embedding. In this work, we propose an efficient dynamic graph embedding approach,  Dynamic Graph Convolutional Network (DyGCN), which is an extension of GCN-based methods. We naturally generalizes the embedding propagation scheme of GCN to dynamic setting in an efficient manner, which is to propagate the change along the graph to update node embeddings. The most affected nodes are first updated, and then their changes are propagated to the further nodes and leads to their update. Extensive experiments conducted on various dynamic graphs demonstrate that our model can update the node embeddings in a time-saving and performance-preserving way. 
\end{abstract}


\maketitle

\section{Introduction} \label{sec:introduction}
\input{1_intro.tex}

\section{Related Works}
\input{2_relatedwork.tex}

\section{Preliminaries}
\input{2_5_preliminaries.tex}

\section{The Proposed Method}
\input{3_method.tex}

\section{Experiments} \label{sec:experiment}
\input{4_experiments.tex}

\section{Conclusions}
\input{5_conclusion.tex}

\bibliographystyle{ACM-Reference-Format}
\bibliography{sample-base}

\end{document}

%% file: 1_intro.tex
Graphs are natural representations for encoding relational structures and therefore have a wide range of applications in bioinformatics, chemistry, recommender systems, social network study, etc. 
By using graphs, researchers obtain rich information from the relation structures, like connectivity, clustering. These rich information is meaningful for many tasks. However, unlike image or language performs in compact grid pattern, graph structure perform in a topological pattern, which is not easy to process for common machine learning method.  Although,  graph could be performed in the form of adjacency matrix, it is too big to save or process, when the graph is big. Graph (network) embedding is considered as the most effective way to represent a graph. Graph embedding aims to find a mapping function which converts each node in the network to a low-dimension latent representation. These representations could infer the characteristic of the entire graph. 

Graph embedding has attracted considerable research efforts \cite{perozzi2014deepwalk,tang2015line,grover2016node2vec}. 
Most researches focus on static graph embedding, i.e., represent nodes of a graph with only the consideration of a static graph. 
There are three popular classes of graph embedding  in recent years. Some of them are based on matrix factorization (MF)~\cite{belkin2002laplacian,cao2015grarep,tang2015line}, typically Belkin et al \cite{belkin2002laplacian} use Singular Value Decomposition (SVD) to decompose the node representation and LINE \cite{tang2015line} applies factorization strategies into large scale graph embedding. 
Some of graph embedding methods are inspired by random walk strategy, including DeepWalk~\cite{perozzi2014deepwalk} and node2vec~\cite{grover2016node2vec}. Random-walk-based methods transform graph structure into a lot of random walk arrays which could be easily processed by skip-gram model in word2vec~\cite{mikolov2013distributed}.
Recently, Graph Convolution Networks (GCNs) \cite{gori2005new,scarselli2008graph,kipf2016semi,hamilton2017inductive,xu2018powerful} as a class of models generalizing neural networks to graph-structured data, have achieved significant attention.
GCN-based methods generally follow an embedding propagation scheme, where the embedding of a node is recursively updated by aggregating message propagated from its neighboring nodes \cite{xu2018powerful}. Under the embedding propagation scheme, GCN-based methods have achieved a great improvement of both efficiency and effective aspects.

Many graphs in real world are inherently dynamic, meaning that their nodes/edges may change over time. 
For instance, in social network, users will join or quit a social network at any time and also they may develop new relationships or break up with others over time. Users may also change their profile information  such as age, location, et al.
Similarly, a citation network is constantly enriched due to the increasing new work citing prior ones and the authors may change their company or interesting area at any time.
Considering graph embedding problem in a dynamic manner is more corresponding to the real-world applications. More information could be captured when we consider the dynamic features of graph.
Although some researches on dynamic graph embedding have been proposed~\cite{tang2008community,sankar2018dynamic,zhou2018dynamic,goyal2018dyngem}, 
they mainly focus on mining the pattern of graph evolvement but ignore the efficiency issue. Most of these works perform traditional static graph embedding in each snapshot of a dynamic graph, and then consider to add constraint or interactions between different snapshots. Their methods consequently get better performance than previous static methods. However, 
generating new representation at each time is costly, since most traditional node representation methods are supposed to learn the embedding parameters by optimization process (e.g., gradient descent or matrix factorization). Repeating these progress at each time step brings about high complexity. 

The time complexity problem mainly comes from two parts. First, global updating time consuming, which means some methods update all the node embeddings at each time step. Second, re-training time consuming, which means some methods update node embeddings by optimization progress, i.e., factorization or Stochastic Gradient Descent.
Both parts of time consuming could be cut down because of the dynamic relations between nearby time steps.
Usually, the change on graphs is frequent and trivial at any time. The node embedding may not change too much, and it is costly and unnecessary to learn the embeddings whole again. 
Some researchers thus try to avoid the re-calculation of all node representation and directly concern obviously changing nodes \cite{hou2019dynwalks,perozzi2014deepwalk}. 
Some researchers avoid the re-learning progress and
directly update the previous embeddings according to the changes~\cite{zhang2018timers,trivedi2018dyrep}. 
Following those ideas, 
researchers consider to design incremental algorithms for traditional graph embedding methods. 
TIMERS~\cite{zhang2018timers} considers the changes as a perturbation of an adjacency matrix and uses matrix disturbance theory to come up with a dynamic graph embedding method based on incremental version of SVD, which does not need re-training process. 
Du et al.~\cite{du2018dynamic} give a dynamic extended version of LINE~\cite{tang2015line} which focuses on how to update the representation of LINE when the graph changes. In this way, the method only needs to update a part of nodes at each time step.
Then DynWalks~\cite{hou2019dynwalks} and NetWalk~\cite{yu2018netwalk} are proposed as the dynamic version of DeepWalk \cite{perozzi2014deepwalk}. They generate new random walks which is highly related the changes of graph at each time step to learn the updating information of corresponding nodes.
As mentioned above, these works extend the static embedding methods to dynamic setting with efficiency. Some focus on cutting down the number of updating nodes at each time step. Some try to avoid the re-learning process such as factorization or Stochastic Gradient Descent.
Recently, GCN shows great power in graph embedding and has the potential to deal with graphs incrementally \cite{hamilton2017inductive}. 
Therefore, we aim to extend GCN to dynamic environment in an efficient manner. 




\noindent \textbf{Present work.}
In this work, we propose an efficient dynamic embedding framework Dynamic Graph Convolutional Network (DyGCN), which is an extension of the GCN-based embedding models.
All methods based on GCN are applicable to our framework.
As its core, we devise a dynamic graph convolution operation which naturally generalizes the embedding propagation scheme of GCN to dynamic setting, which is to propagate the change along the graph to update node embeddings.  
In detail, it consists of two operations: calculating the change of message aggregated from neighbors and adjusting the node embeddings according to it.
We first update the most affected nodes, and then their change is propagated to the further nodes and leads to their update. We design two different ways for node propagation, high-order update and spectral propagation.
Comparing with high-order update and spectral propagation, the the former
is more efficient but the latter performs more precisely. Both ways cut down the computational time compared with other methods of dynamic graph embedding.
We perform extensive experiments on three real-world datasets.  
Comparison with other methods and the static counterpart (GCN) demonstrates the efficiency and effectiveness of the proposed method. 
Overall, our contributions can be summarized as follows: 
\begin{itemize} 
  \item We propose DyGCN, an efficient dynamic embedding framework for GCN-based embedding methods, which is able to update node embeddings in dynamic graphs in a time-saving and performance-preserving way.
  \item As the core of GCN, we devise a dynamic graph convolution operation which propagates the change along the graph to update node embeddings, which can avoid the costly re-learning progress.
  \item Extensive experiments conducted on various dynamic graphs demonstrate that the effectiveness and efficiency of our approach.
\end{itemize}     

%% file: 2_relatedwork.tex
In this section, we briefly review three streams of research related to our work: network embedding methods, dynamic embedding methods and graph neural networks. Note that, we only discuss on unsupervised network embedding methods in this paper.

\subsection{Network embedding methods}
Network embedding is an important research area for graph analysis. From the beginning, network embedding is described as a typical task of dimension reduction problem. It is defined as representing a $n \times n$ adjacency matrix as a $n \times k$ matrix, where $k \ll n$. 
Traditional dimension reduction methods are applied, such as Principal Component Analysis (PCA) \cite{wold1987principal} and Multi-Dimensional Scaling (MDS) \cite{kruskal1978multidimensional}. These methods are generally used in dimension reduction area and also performs well in network embedding. In early years, researchers draw more attention on the specific characteristics of networks. Then, some methods such as ISOmap \cite{tenenbaum2000global}, and Local Linear Embeddings (LLE) \cite{roweis2000nonlinear} are designed for better global structure preservation. Meanwhile, another class of dimension reduction methods become popular in networks area. They use the spectral properties of graph to embed networks in spectral domain, such as Laplacian Eigenmaps (LE) \cite{belkin2002laplacian}. These methods control the global or local information that need to preserved by filtering different sizes of eigenvalues. 
These methods above all suffer from computation complexity problem. To solving this problem, stochastic gradient descent based factorization methods are commonly used to apply network embedding to lager scale dataset, such as GraRep \cite{cao2015grarep}, LINE \cite{tang2015line}. 

In recent years, deep learning methods have already achieved better results in network embedding areas. Different from traditional methods which consider network embedding as dimension reduction, deep learning methods consider it as a representation learning problem. The first deep learning method of network embedding is DeepWalk \cite{perozzi2014deepwalk}. DeepWalk transforms graph structure information into sequences by random walk, and then Skip-Gram \cite{mikolov2013efficient} is applied on it for each node embedding. After the success of DeepWalk, more deep learning methods are designed in network embeddings. Node2vec \cite{grover2016node2vec} improves the random walk strategies of DeepWalk by a controllable deep or wide walking possibility. SDNE \cite{Wang2016Structural} introduces Deep Autoencoder model to network embedding, modeling 1st-order and 2nd-order neighbors of nodes. Nowadays, graph neural network catches a lot attention for its ability of modeling complex graph relations, which highly fit the request of network embedding. For example,  VGAE \cite{kipf2016variational} and GraphSAGE use graph convolution network to model information propagation in graph, 
GAT \cite{velivckovic2017graph} applied attention mechanism to modeling the relations between neighbors.

\subsection{Dynamic Graph Embedding Methods} \label{sec:related_work_dge}
There are two streams of researches on dynamic graphs. One aims to \emph{predict} the graph structure at the next time \cite{seo2018structured,yu2017spatio,pareja2019evolvegcn}, called dynamic graph prediction, the other tries to \emph{learn} the better graph representation according to dynamic information, called dynamic graph embedding, which is mainly discussed in our paper. In that case, we don't compare DyGCN with some well-known graph prediction methods such as EvolveGCN \cite{pareja2019evolvegcn}, DynGraph2Vec \cite{goyal2020dyngraph2vec}. 


At beginning, Dynamic graph embedding is considered as an extension of network embedding. Researchers simply use static network embedding methods in each  snapshot during discrete time steps, with constraint to align the same nodes in different time steps. 
In this way, researchers claim that dynamic graph embedding methods could obtain better representations than traditional static network embedding, since different snapshots share more characteristics for representation. Zhou et al~\cite{zhou2018dynamic} constraints the representation in each time step, by the formulation that triadic closure process is more frequent along graph evolving.
Zuo et al~\cite{zuo2018embedding} applies a Hawkes process to model the evolution of neighborhood formation sequence.
Goyal et al~\cite{goyal2018dyngem,goyal2020dyngraph2vec} use Auto-Encoder to learn the graph and use the Recurrent Neural Network (RNN) to model the relations over time.
Sankar et al~\cite{sankar2018dynamic} applies attention mechanism in it, learning structural and temporal attention to adaptively obtain useful information for embedding.

An inevitable problem comes from above methods. Although they achieve better representation than static methods, their computation complexity is unacceptable for real-time application. Since these methods rely on the static representation of every snapshot, they are supposed to perform static network embedding process every time step at least, which is difficult to use online.
For the sake of efficiency, researchers try to update some embeddings of previous snapshots without re-calculating the current graph totally. 
TIMERS \cite{zhang2018timers} proposes an incremental Singular Value Decomposition (SVD) model for dynamic embedding, which only needs SVD  at beginning, and incrementally updates them according to graph changing. 
DNE \cite{du2018dynamic} proposes an dynamic version of LINE, with only a few gradient descent process to update the representation of current graph.
DANE \cite{li2017attributed} proposes a network embedding method in dynamic environment which updates the node embedding based on the change on adjacency matrix as well as attribute matrix via matrix perturbation.
DyRep \cite{trivedi2018dyrep} ingests dynamic graph information in the form of association and communication events over time and updates the node representations as they appear in these events. 
DynWalks \cite{hou2019dynwalks} incorporates the temporal information with traditional DeepWalk to capture the evolving properties in dynamic networks. It updates nodes embeddings by sampling new walks which is highly related to the changes of graph.

Dynamic graph prediction task is different from dynamic graph embedding, but there are still some similarities. To avoid misunderstanding, we also list some typical methods of this class below, which would not be discussed in this paper. 
GCRN \cite{seo2018structured} uses a Graph Convolutional Networks (GCN) to learn node embedding and feed it into the LSTM to learn dynamism.
WD-GCN/CD-GCN \cite{manessi2019dynamic} modifies the LSTM by incorporating with GCN.
STGCN \cite{yu2017spatio} proposes a so-called ST-Conv blocks, which requires that the node features must be evolving over time.
EvolveGCN \cite{pareja2019evolvegcn} utilizes RNN to evolve the GCN parameters, instead of the node embedding.
In summary, these aforementioned work mostly utilize GCN to generate node embeddings and RNN to handle dynamism, which aims to model the evolving patterns and prediction.
In contrast, our model is designed to extend GCN to efficiently learn the node embeddings in dynamic environment.

\subsection{Graph neural networks}
Graph neural networks (GNN), extending deep neural networks to graph structured data, are a class of models recently advance in graph representation learning.
Due to convincing performance and high interpretability, GNN has been widely applied on many applications like neural machine translation \cite{beck2018graph}, image classification \cite{marino2017more}, recommendation \cite{Wu2018Session} and fashion analysis \cite{cui2019dressing}.
The concept of GNN is first proposed by \cite{scarselli2009graph}.
Most GNN models submit to similar scheme.
Generally, nodes in GNNs aggregate information from neighborhoods and update their embeddings iteratively.                     
There have been various kinds of GNNs proposed these days.
Comprehensive surveys can be found in \cite{wu2019comprehensive,zhou2018graph}.
Graph Convolutional Networks (GCN) \cite{kipf2016semi} perform graph convolutions for aggregation and update motivated by spectral convolution.
Gated Graph Neural Networks (GGNN) \cite{li2015gated} aggregates and updates in the form of Gated Recurrent Unit (GRU).
Graph attention network (GAT) \cite{velivckovic2017graph} incorporates the attention mechanism into the aggregation step.
GraphSAGE \cite{hamilton2017inductive} considers from the spatial perspective and introduces an inductive learning method. 
After that, researchers focus on the effectiveness and efficiency of aggregation and updating part. 
GIN \cite{xu2018powerful} first attempts to study the representation ability of different kinds of aggregation part, and gets a more powerful GNN model. 
FastGCN \cite{chen2018fastgcn} deals with the computation complexity problem of GCN, it samples node by each layer of aggregation.
SGC \cite{wu2019simplifying} claims that the nonlinear part of GCN layer, i.e., activate function, is meaningless. They remove activate functions between GCN layers and only remain the nonlinear layer in classifier, and achieve comparable results as common GCN methods.


%% file: 2_5_preliminaries.tex
In this section, we first introduce the traditional graph convolutional network in detail. Then, we formulate the graph embedding problem from static and dynamic styles.

\subsection{Static graph embedding}
Given a graph $\mathcal{G}=\left \{ \mathcal{V}, \mathcal{E} \right \}$, where $\mathcal{V} = [v_{1}, v_{2}, ..., v_{N}]$ and $ \mathcal{E} = [e_{1}, e_{2}, ..., e_{M}]$ denotes the node and edge set in $\mathcal{G}$ respectively. $N$ is the number of nodes and $M$ is the number of edges. We denote the attributes of each node as $X$, where $X_{v}$ represents the attribute of node $v$ in $\mathcal{V}$. The attribute information includes many aspects, such as node degree and other information according to different kinds of datasets.  The adjacency matrix is defined as $A$.  The  static graph embedding methods use $A$ (or $A, X$) to obtain the node embedding matrix $Z \in \mathbb{R}^{N \times d}$, 
\begin{equation}
\label{equ:static_GE}
Z = \text{Static\_GE}(A, X)
\end{equation}
where $d$ is the dimension of the embeddings. Each row of $Z$ is $z_{v}$ denotes the representation of node $v$. The purpose of static  graph embedding methods is to obtain better $z_{v}$  which could express both the structural and attributive feature of node $v$ for down steam tasks such as node classification and link prediction.  


\subsection{Dynamic graph embedding}
Different from  static graph embedding problem, dynamic graph embedding extend the formulation by time step. 
Henceforth we use the superscript $t$ to denote the time index.
In that case, we have a graph sequence as $[\mathcal{G}^{1}, \mathcal{G}^{2}, ..., \mathcal{G}^{t}, ..., \mathcal{G}^{T}]$, where $\mathcal{G}^{t}=\left \{ \mathcal{V}^{t}, \mathcal{E}^{t} \right \}$ represent the graph at time step $t$. $T$ is the length of graph sequence. $\mathcal{V}^{t}$ and $\mathcal{E}^{t}$ are the set of nodes and edges in $\mathcal{G}^{t}$ respectively.  Without loss of generality, we assume that the graph at any time is built on a common node set of cardinality $N$.  The nonexistent node is treated as a dangling node with zero degree. If there is a node added/removed, we add/remove the corresponding edges. In this case, the changes of nodes can also be seen as those of edges. In the following, we only talk about the changes of edges.


The node feature matrix $X$ is commonly stable during dynamic graph. In this paper, we mainly discuss the evolving adjacency matrix $A^{t}=\left \{ 0,1 \right \}^{N \times N}$.
$Z^{t}\in \mathbb{R}^{N \times d}$ is the node embeddings matrix, each row of which is the $d$-dimensional embedding of a node in $\mathcal{V}^{t}$. Dynamic graph embedding methods could be formulated as, 
\begin{equation}
\label{equ:dynamic_GE}
Z^{t} = \text{Dyn\_GE}(A^{\leq t}, X), 
\end{equation}
where $A^{\leq t}$ means the adjacency matrices from the beginning to time $t$. 

In dynamic graph embedding, there are two different research topics. The first one focus on get more representative embeddings by considering  previous graphs information. Theoretically, the lower bound of the representation capacity  is the static graph embedding representation.  The second one focuses on the efficiency of dynamic graph representation. In that case, the static graph embedding representation is the upper bound of the representation capacity. 
The main idea for cutting down the computation is to update the previous node embeddings by $\Delta A^{t}$ without learning the representation at each time, which is formulated as, 
\begin{equation}
\label{equ:dynamic_geeff}
Z^{t} = \text{Dyn\_GE\_efficient}(\Delta A^{t-1}, Z^{t-1}), 
\end{equation}
where the initial matrix embedding  $Z^{0}$ is obtained by static graph embedding method according to different dynamic graph methods. $\Delta A^{t-1} = A^{t} - A^{t-1}$ is the changing of graph.

\subsection{Graph Convolutional Network} \label{sect:gcn}
Graph Convolutional Network (GCN) \cite{kipf2016semi,hamilton2017inductive,velivckovic2017graph} is a class of models, which extend neural networks to graphs. 
It takes the adjacency matrix $A$ and node features matrix $X$ as input and output the node embedding matrix $Z \in \mathbb{R}^{n \times d}$, where $n$ is the number of nodes and $d$ is the dimensionality of node embeddings:
\begin{equation} \label{eq:gcn}
Z = \text{GCN}(A,X).
\end{equation}
A GCN consists of multiple layers of graph convolution to iteratively update the node embeddings. 
Specifically, the node embeddings $Z$ can be update to $Z^{'}$ via a layer of graph convolution as (omitting the layer index for clarity):
\begin{equation} \label{eq:gconv}
\begin{aligned}
Z^{'} & = \text{GCONV}(A, Z, W) \\
& :=\sigma(\hat{A}ZW).
\end{aligned}
\end{equation}
$\hat{A}$ is some normalization of the graph adjacency matrix. 
We choose $\hat{A}=A+I$ in this paper for simplicity 
, where $I$ is the identity matrix.
$W$ is a layer-specific transformation matrix and $\sigma$ is a nonlinear activation such as ReLU.

The graph convolution is actually an embedding propagation operation motivated by spectral convolution.
Here we reformulate Equation \ref{eq:gconv} from the perspective of a single node $v$, which is totally the same operation as the former Equation.
For notional clarity, we write the embedding of a node $v$ as $z_{v}$, which is a row vector of $Z$. 
A node $z$ first aggregates information propagated from neighbors and itself.
\begin{equation} \label{eq:aggr} a_{v} = \sum_{u \in \mathcal{N}(v)\cup v}z_{u},
\end{equation}
where $a_{v}$ is the aggregated information of node $v$ and $\mathcal{N}(v)$ is the set of its neighbors.
Then it generates a new embedding $z^{'}_{v}$ based on the aggregated information as,
\begin{equation} \label{eq:update}
z^{'}_{v} = \sigma(a_{v}W).
\end{equation} 
The initial node embeddings are input node features $X$.
After limited numbers of graph convolution, the node embeddings will capture the information propagated from various orders away and output as the final node embedding $Z$.

%% file: 3_method.tex
\begin{figure*}[t]
\centering
\includegraphics[width=1\linewidth]{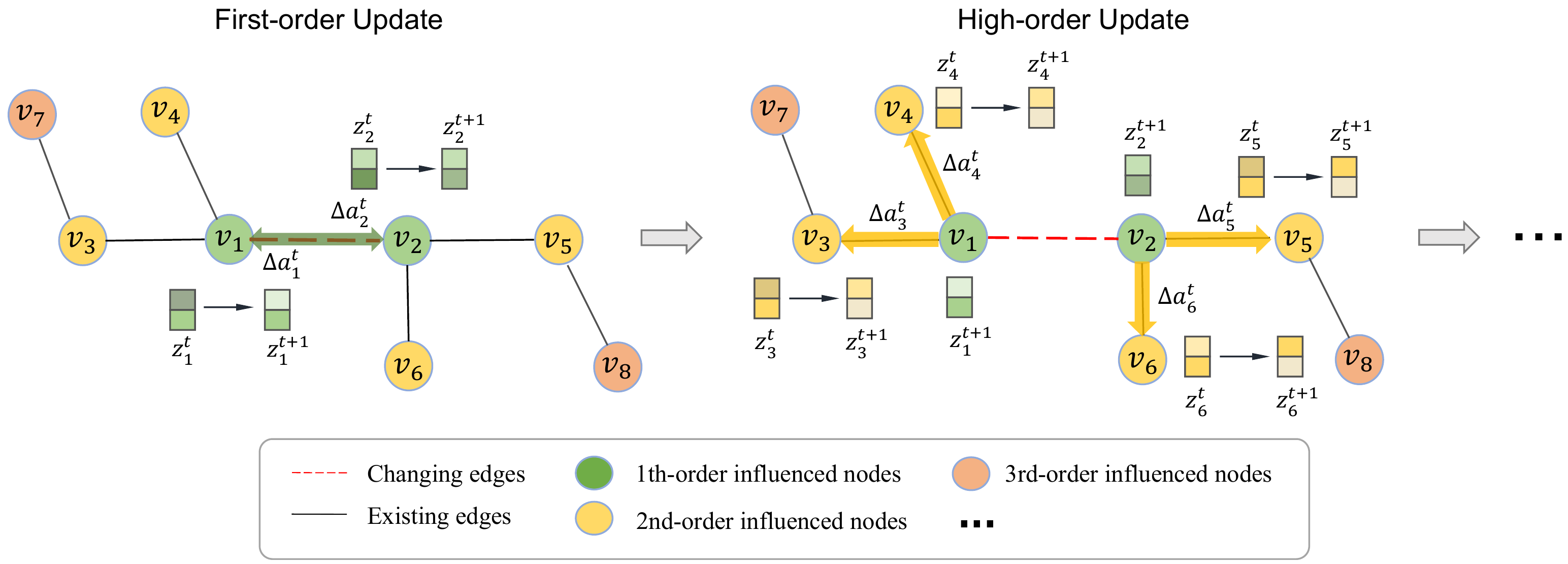} 
\caption{The overview of DyGCN. For an edge emerging at time $t$, $v_{1}$ and $v_{2}$ are the nodes directly influenced by it, which are first updated via the first-order propagation.
Then the updated $v_{1}$ and $v_{2}$ affect their neighbors $v_{3}$, $v_{4}$, and $v_{5}$, $v_{6}$, which are updated accordingly to the changed aggregated message.
Analogously, the further nodes are updated in turn.}
\label{fig:overview}
\end{figure*}

In this section, we first present a naive graph convolution network for graph embedding. Then, we introduce two versions of  our proposed model  DyGCN extending static GCN to dynamic setting. Finally, we discuss our methods with other graph models from effectiveness and efficiency aspects.

\subsection{Naive Graph Convolutional network for Graph Embedding}


A naive way to utilize GCN in this dynamic setting is to directly apply it at each snapshot to learn the new node embeddings as
\begin{equation} \label{eq:gcn1}
Z^{t+1} = \text{GCN}(A^{t+1},X).
\end{equation}
Nevertheless, it is costly and unnecessary since the change is usually trivial at each time.
In particular, there is little difference between $A^{t+1}$ and $A^{t}$, or in other words,  $\Delta A^{t}$ is sparse.

\subsection{Dynamic Graph Convolutional Network}
Intuitively, the node representation  tends to get the information of the another node when their link emerges. For example in recommendation system, when a user interacts with baby carriage
, the user may intend to get more preference on baby care in future.
In this subsection, we elaborate the proposed model DyGCN, whose framework is shown in Figure \ref{fig:overview}.
We first introduce some notations. 
An example of a dynamic graph is shown in the left side, where an edge between node $v_{1}$ and $v_{2}$ emerges at time $t$.  
We will propagate the change along the graph to update node embeddings in turn.
$v_{1}$ and $v_{2}$ are denoted as the 1st-order influenced nodes which are directly influenced by the emerging edge.
$v_{3}$ and $v_{4}$, $v_{5}$ and $v_{6}$, as the neighbors of the 1st-order influenced nodes $v_{1}$ and $v_{2}$, will be affected. We denote them as the 2nd-order influenced nodes.
Analogously, $k$th-order influenced nodes refer to the nodes that are $(k-1)$ hops away from 1st-order influenced nodes.
We denote the set of $k$th-order influenced nodes at time step $t$ as $V_{k}^t$. 

Our devised model is based on the embedding propagation scheme of GCN~\cite{kipf2016semi,hamilton2017inductive,xu2018powerful}.
The node embeddings are updated according to the change of aggregated message.
In the first-order propagation layer, we update 1st-order influenced nodes, which are the nodes directly connected to the changing edges.
Then the updated embeddings are propagated to their neighbors and lead to their update in the successive high-order propagation layers.
In the following, we illustrate their design in turn.

\noindent\textbf{First-order Update.~}
Here we talk about how to update embeddings of 1st-order influenced nodes.
Let us first review the aggregation and update scheme of GCN in Eq \ref{eq:aggr} and \ref{eq:update}. 
A node first aggregates message propagated from neighbors and itself by summation, and then update embeddings based on it.
It is the aggregated message that decides on the final embeddings.
The change of topological structure will influence the aggregated message and thus the final node embeddings.

Formally, for a 1st-order influenced node $v \in V_{1}^{t}$, the change of its aggregated information at time $t$ can be calculated as\footnote{For notional clarity, we use $z$ to denote node embedding in dynamic setting, distinguished from $h$ in the Section of Graph Convolutional Network.},
\begin{equation} \label{eq:change_aggr1}
\Delta a_{v}^{t} = \sum_{u \in \mathcal{N}^{t+1}(v)\cup v}z_{u}^{t}-\sum_{u \in \mathcal{N}^{t}(v)\cup v}z_{u}^{t},
\end{equation}
where $z_{v}^{t}$ is the embedding of node $v$ at time step $t$.
$\mathcal{N}^{t}(v)$ and $\mathcal{N}^{t+1}(v)$ denote the neighbors of node $v$ at time $t$ and $t+1$, respectively.
It is actually the summation of the embeddings of newly appearing neighbors subtracts that of the disappearing neighbors.
For example, $v_{1}$ in Figure \ref{fig:overview} has a new neighbors $v_{2}$, the change of its aggregated message is thus the embedding of node $v_{2}$, i.e., $\Delta a_{1}^{t}=z_{2}$.
If the edge between $v_{1}$ and $v_{2}$ is removed instead, then $v_{1}$ loses the neighbor $v_{2}$, in which case the change of its aggregated message $\Delta a_{1}^{t}=-z_{2}$.

The original graph convolution generates   the node embedding based on the aggregated message via a transformation matrix.
Drawing lesson from that, we introduce an extra transformation matrix $W_{1}\in \mathbb{R}^{d \times d}$ to model the influence of the change of aggregated message on 1st-order influenced nodes' embeddings.
Specifically, it's a neural network based function to update the node embedding based on the original node embedding and the calculated change of aggregated message as,
\begin{equation} \label{eq:1update}
z_{v}^{t+1} = \sigma(W_{0}z_{v}^{t}+W_{1}\Delta a_{v}^{t}),
\end{equation}
where $W_{0}\in \mathbb{R}^{d \times d}$ is a transformation matrix for original embeddings.

\noindent\textbf{High-order Update.~}
Since the 1st-order influenced nodes have been updated, their changed embeddings will be propagated to their neighbors and influence their embeddings.
We here introduce the update of 2nd-order influenced nodes, which can be easily generalized to the further neighbors.
 
In specific, the aggregated message of 2nd-order influenced nodes alters since the embeddings of some neighbors (the 1st-order influenced nodes) has changed.
Different from the 1st-order influenced nodes influenced by the change of topological structure, the 2nd-order influenced node is influenced by the change of their neighbors' embeddings.
Formally, we can calculate the change of aggregated message of a 2nd-order influenced node $v \in  V_{2}^{t}$ as:
\begin{equation} \label{eq:change_aggr2}
\Delta a_{v}^{t} = \sum_{u \in \mathcal{N}^{t+1}(v)\cup v}(z_{u}^{t+1}-z_{u}^{t}).
\end{equation}
In essence, it is only the change of the 1st-order influenced nodes in their neighborhood since the others' embeddings remain unchanged.
Back to the example in Figure \ref{fig:overview}, the node $v_{3}$ aggregates the embeddings of its neighbor, the 1st-order influenced node $v_{1}$ whose embedding has been updated to $z_{1}^{t+1}$.
Accordingly, the change of its aggregated message is $z_{1}^{t+1}-z_{1}^{t}$.

We use a similar update function as in Equation \ref{eq:1update} to update the embeddings of a 2nd-order influenced node $v$ with another transformation matrix $W_{2}\in \mathbb{R}^{d \times d}$:
\begin{equation} \label{eq:2update}
z_{v}^{t+1} = \sigma(W_{0}z_{v}^{t}+W_{2}\Delta a_{v}^{t}).
\end{equation}

Analogously, the update of 2nd-order influenced nodes will be propagated to their neighbors iteratively, which can also be updated in this way.
Formally, for a $k$th-order influenced nodes $v \in \mathcal{V}^t_k$, they aggregate changed message from ($k$-1)th-order influenced nodes as in Eq~\ref{eq:change_aggr2} and update itself:
\begin{equation} \label{eq:kupdate}
z_{v}^{t+1} = \sigma(W_{0}z_{v}^{t}+W_{k}\Delta a_{v}^{t}),
\end{equation}
where $W_{k}\in \mathbb{R}^{d \times d}$ is the transformation matrix for $k$-th propagation layer.
Overall, the algorithm of DyGCN can be summarized in Algorithm \ref{alg:dygcn}.

  \begin{algorithm}[H]
    \caption{DyGCN} 
    \label{alg:dygcn} 
    \begin{algorithmic}[1] 
    \REQUIRE
    $\mathcal{G}^{t}=\left \{ \mathcal{V}^{t}, \mathcal{E}^{t} \right \}, G^{t+1}=\left \{ \mathcal{V}^{t+1}, \mathcal{E}^{t+1} \right \}$: the graph at time $t$ and $t+1$; \\
    $Z_{t}=\left \{ z_{v}^{t}, v \in \mathcal{V}^{t} \right \}$: the node embeddings at time $t$;  \\
    $\left \{W_{0}, W_{1}, W_{2}, ..., W_{K}\right \}$: the transformation matrices.\\
    \ENSURE 
    $Z^{t+1}=\left \{ z_{v}^{t+1}, v \in \mathcal{V}^{t} \right \}$, the node embeddings at time $t+1$. \\
\STATE $//$ The update of first-order influenced nodes \\ 
\FOR{$v \in \mathcal{V}^{t}_{1}$}
\STATE $\Delta a_{v}^{t} = \sum_{u \in \mathcal{N}^{t+1}(v)\cup v}z_{u}^{t}-\sum_{u \in \mathcal{N}^{t}(v)\cup v}z_{u}^{t};$ \\
\STATE $z_{v}^{t+1} = \sigma(W_{0}z_{v}^{t}+W_{1}\Delta a_{v}^{t})$; \\
\ENDFOR
\STATE  $//$ The update of high-order influenced nodes \\
\FOR{$k \in \left [2, ..., K  \right ]$}
\FOR{$v \in \mathcal{V}^{t}_{k}$}
\STATE $\Delta a_{v}^{t} = \sum_{u \in \mathcal{N}^{t+1}(v)\cup v}(z_{u}^{t+1}-z_{u}^{t});$ \\
\STATE $z_{v}^{t+1} = \sigma(W_{0}z_{v}^{t}+W_{k}\Delta a_{v}^{t})$.
\ENDFOR
\ENDFOR
\end{algorithmic}
\end{algorithm}

\noindent\textbf{The Process of New Nodes.~}
Our method mainly discuss the structural changes i.e., the changing of edges. The new nodes emerging at time $t$ could be considered as some isolated nodes before time $t$. In that case, we define the $A^{0}$ large enough for adding nodes though all the time steps. Our method could deal with new nodes without any changes of the formulation above.
In the other word, for each new node at time step $t$, we initialize the embeddings as the aggregation of the its linked nodes at time step $t$.

\noindent\textbf{Matrix Form.~}
Above we detail the update process from the perspective of embedding propagation scheme.
To offer a holistic view of embedding propagation and facilitate batch implementation, we here provide the matrix form of the update rule as in GCN \cite{kipf2016semi}. which is denoted as dynamic graph convolution.
An extra subscript is introduced to denote the layer index of dynamic graph convolution in DyGCN.
The above discussed 1st-order update can be summarized in the following expression, called 1st-order dynamic graph convolution:
\begin{equation} \label{eq:dyn_gcn1}
\begin{aligned}
Z^{t+1}_{1} & = \text{DYGCONV}_{1}(\Delta A^{t}, Z^{t}, W_{0}, W_{1}) \\
& :=\sigma(\Delta \hat{A}^{t}Z^{t}W_{1}+Z^{t}W_{0}),
\end{aligned}
\end{equation}
where $Z^{t+1}_{1}$ denotes the node embeddings after 1st-order update.
Since we only aim to update the 1st-order influenced nodes here, we mask all the other nodes in the above update process.
 
For the $k$-order influenced nodes ($k>2$), their update can be written as Equation (\ref{eq:dgc}), called $k$-order dynamic graph convolution:
\begin{equation} \label{eq:dgc}
\begin{aligned}
Z^{t+1}_{k} & = \text{DYGCONV}_{k}(A^{t+1}, \Delta Z^{t}, Z^{t},  W_{0}, W_{k}) \\
& :=\sigma(\hat{A}^{t+1}\Delta Z^{t}W_{k}+Z^{t}W_{0}),
\end{aligned}
\end{equation}
where $\Delta Z^{t}=Z^{t+1}_{k-1}-Z^{t+1}_{k-2}$.
$Z^{t+1}_{k}$ denotes the node embeddings after $k$-order updated.
Likewise, we mask all the nodes except $k$-order influenced nodes in the above update process.
After $K$ times of dynamic graph convolution, the final node embeddings $Z^{t+1}$ is obtained by updating all the k-order nodes with corresponding column of $Z_{k}^{t+1}$.

\noindent\textbf{Unsupervised Training.~}
In order to learn meaningful and predictive representations in a fully unsupervised setting, we apply a structure-preserving loss function on the updated node embeddings to tune the transformation matrices $\left \{ W_{0}, W_{1}, \cdots, W_{K} \right \}$.  
The training dataset is generated as a sequence of graph. We use a trained static graph embedding methods to get the $Z^{0}$ and reconstructing the graph at other times by a structure-preserving loss function as the training process.
The structure-preserving loss function encourages nearby nodes to have similar embeddings, while enforcing that the embeddings of disparate nodes are highly distinct \cite{hamilton2017inductive}:
\begin{equation}\label{eq:neg}
L = -\sum_{t} \sum_{v} \left(\log\left(\sigma(z_{v}^{\top}z_{u}) \right)+ \log\left(1 -\sigma(z_{v}^{\top}z_{u^{-}}) \right)\right)
\end{equation}
where $u$ is a neighbor of $v$ and $u^{-}$ is a random node. $\sigma$ is the sigmoid function. 
The unsupervised loss aims to preserve the topological information, which requires no extra supervision. 
It can be replaced, or augmented, by a task-specific objective (e.g., cross-entropy loss) for specific downstream tasks.

\subsection{Spectral DyGCN}
In this section, we introduce Spectral DyGCN to further improve the high-order update mechanism above with acceptable time consumption.

The high-order update mechanism ignores the global propagation of changing nodes. The high-order update mechanism described above have to ignore the nodes with order higher than $k$. It means that the changing information of 1-order nodes could not be propagated to all the nodes, which inevitably brings about the loss of accuracy. However, the influence from 1st-order nodes to the graph globally. Every node is affected by its neighbors, not simply from 1st-order nodes to 2nd-order nodes and 2nd-order nodes to 3rd-order nodes.

 The global propagation of graph is difficult in spatial form of GCN, since spatial convolution originally consider each node separately. Although it is not easy to tell which node influences more in spatial domain.  we consider the graph propagating step in spectral domain.
 In graph theory, the normalized graph Laplacian is defined as $ L= I_{n} - D^{-\frac{1}{2}} A D^{-\frac{1}{2}}$, where $I_{n}$ is the identity matrix.
 The normalized Laplacian can be decomposed as $L=U \Lambda U^{-1}$, where $\Lambda=diag([\lambda_{1}, ..., \lambda_{n}])$ as its eigenvalues and $U$ is the $n \times n$ square matrix whose
$i$th column is the eigenvector $u_{i}$. The graph Fourier transform of a signal $x$ is defined as $\hat{z} = U^{-1}z$, while the inverse transform is $x = U\hat{x}$. Then the original network propagation $D^{-\frac{1}{2}} A D^{-\frac{1}{2}} z$ could be interpreted that $x$ is first transformed into the spectral space and scaled by the eigenvalues, and then transformed back is written as,  \ref{eq:porpagation}.
\begin{equation} \label{eq:porpagation}
z^{'} = D^{-\frac{1}{2}} A D^{-\frac{1}{2}} z = U (I_{n} - \Lambda) U^{-1} z. 
\end{equation}
It has been proved that the scale of eigenvalues control how the propagation process, partially or modularly \cite{zhang2019prone}. We generalizes the network propagation of Equation \ref{eq:porpagation} into a learnable form for better approximation of dynamic propagation,
\begin{equation} \label{eq:dyn_porpagation}
z^{'} = U g_{\theta}(\Lambda) U^{-1} z.
\end{equation}
As is proved in \cite{kipf2016semi}, Equation \ref{eq:dyn_porpagation} could be  well-approximated by a truncated expansion in terms of Chebyshev polynomials and the  graph propagating step is formulated as,
\begin{equation} \label{eq:spectural_dygcn}
Z^{'} \approx  W_{s} (I_{N}+D^{-\frac{1}{2}} A D^{-\frac{1}{2}}) Z,
\end{equation}
where $W_{s}$ is the parameter matrix after approximation.
Equation \ref{eq:spectural_dygcn} is exactly the formulation of vanilla GCN model in spectral domain. 
 In that case, Spectral DyGCN mainly changes the high order update part of DyGCN to propagate the changing information to the whole graph, and the algorithm is listed as Algorithm \ref{alg:spectral_dygcn}. 

  \begin{algorithm}[H]
    \caption{Spectral DyGCN} 
    \label{alg:spectral_dygcn} 
    \begin{algorithmic}
    
    \REQUIRE
    $\mathcal{G}^{t}=\left \{ \mathcal{V}^{t}, \mathcal{E}^{t} \right \}, \mathcal{G}^{t+1}=\left \{ \mathcal{V}^{t+1}, \mathcal{E}^{t+1} \right \}$: the graph at time $t$ and $t+1$; \\
    $Z_{t}=\left \{ z_{v}^{t}, v \in \mathcal{V}^{t} \right \}$: the node embeddings at time $t$;  \\
    $\left \{W_{0}, W_{1}, W_{2}, ..., W_{K}\right \}$: the transformation matrices.\\
    \ENSURE
    $Z^{t+1}=\left \{ z_{v}^{t+1}, v \in \mathcal{V}^{t} \right \}$, the node embeddings at time $t+1$.
    
\STATE $//$ The update of first-order influenced nodes \\ 
\FOR{$v \in \mathcal{V}^{t}_{1}$}
\STATE $\Delta a_{v}^{t} = \sum_{u \in \mathcal{N}^{t+1}(v)\cup v}z_{u}^{t}-\sum_{u \in \mathcal{N}^{t}(v)\cup v}z_{u}^{t};$ \\
\STATE $z_{v}^{t+1} = \sigma(W_{0}z_{v}^{t}+W_{1}\Delta a_{v}^{t})$  \\
\ENDFOR
\STATE $//$ The propagating the changes to the whole graph \\
\STATE $Z^{t+1} = W_{s} (I_{N}+D^{-\frac{1}{2}} A D^{-\frac{1}{2}}) Z^{t+1}$
\end{algorithmic}

\end{algorithm}

\subsection{Model Analysis}
In this section, we mainly compare DyGCN with other models from effectiveness and efficiency aspects. The demonstrative result could be found in Section \ref{sec:experiment}.
There are mainly two parts of time consuming problems in common dynamic graph embedding, global update consumption and re-training consumption. We mainly summarize them on different compared methods on Table \ref{tab:time_analysis}.

\begin{table}[t]
\centering\caption{Time consuming analysis on different compared methods. $\lvert \mathcal{E}^{t}\rvert$ and $\lvert \mathcal{V}^{t}\rvert$ mean the number of edges and nodes of graph. $\lvert  \Delta \mathcal{E}^{t}\rvert$ and $\lvert  \Delta \mathcal{V}^{t}\rvert$ denote the number of  changing edges and nodes. $d$ is the dimensionality of node embeddings. $r$, $l$, $w$ are parameters of RandmWalk-based methods, i.e.,  the number of walks, walk length and the window size correspondingly. }
\begin{tabular}{|c|c|c|} 
\hline
Methods & global updating   & re-training \\
\hline
GCN & $O(d \lvert \mathcal{E}^{t}\rvert )$ & \Checkmark \\
DNE & $O(d \lvert \Delta \mathcal{E}^{t}\rvert )$  & \Checkmark \\
DynWalks &  $O(\lvert \Delta \mathcal{E}^{t}\rvert+ \lvert \mathcal{V}^{t}\rvert + rwl \lvert \Delta \mathcal{V}^{t}\rvert )$ & \Checkmark \\
GraphSAGE & $O(d \lvert \Delta \mathcal{V}^{t}\rvert )$ & \XSolidBrush  \\
DyGCN & $O(d (\lvert \Delta \mathcal{E}^{t}\rvert + \lvert \Delta \mathcal{V}^{t}\rvert ))$ & \XSolidBrush   \\
Spectral DyGCN & $O(d (\lvert \Delta \mathcal{E}^{t}\rvert + \lvert \mathcal{V}^{t}\rvert ))$ & \XSolidBrush   \\
\hline
\end{tabular}\label{tab:time_analysis}
\end{table}

\noindent\textbf{DyGCN VS static methods (e.g. GCN).~}
The vanilla GCN designed for static setting consists of multiple layers of graph convolution, which update the node embeddings between two layers.
In contrast, our proposed DyGCN is designed for dynamic setting consists of multiple layers of dynamic graph convolution to update the node embeddings between two consecutive time steps.
The 1st-order and high-order update can be approximately seen as graph convolutions on the change of adjacency matrix $\Delta A^{t}$ and $\Delta Z^{t}$, respectively.
The time complexity of graph convolution comes from the matrix multiplication of adjacency matrix, node embedding matrix and transformation matrix.
Satisfyingly, there is mostly $0$ in $\Delta A^{t}$, and the number of non-zero entries in $\Delta A^{t}$ equals to the number of 1st-order influenced nodes.
Likewise, the number of non-zero entries in $\Delta Z^{t}$ at $k$-order update equals to the number of $k$-order influenced nodes.
Compared with static GCN that updates the embeddings of all the nodes, what we update $V_1^t \cup \cdots \cup V_K^t$ is a subset of the whole node set $\mathcal{V}^t$.
Considering $K$ is usually small (2 by default), the number of updated nodes is much less than the size of the whole node set and thus much running time is saved.
In addition, the vanilla GCN is supposed to learn the parameters for each graph for better convolution process, which means it needs to train the model at every time step, which is also a big computational problem.

\noindent\textbf{DyGCN VS DNE and DynWalks.~}
DNE~\cite{du2018dynamic} and DynWalks~\cite{hou2019dynwalks} are dynamic extended version for LINE and DeepWalk respectively. They both consider how to update only a few nodes according to the changes, which means their mainly solve the global update consumption problem.
However, both of these methods update node embeddings by stochastic gradient descent. which means the optimization progress during update is necessary. In contrast, DyGCN does not need optimization progress once the model has been trained, which make it much more time-saving.

\noindent\textbf{DyGCN VS GraphSAGE.~}
Different from vanilla GCN mentioned above, GraphSAGE solves both time consuming problems. First, GraphSAGE is designed to learn the interaction mechanism between different nodes for a type of graph. When there come new nodes, the model does not need to be re-trained. Since the interaction mechanism is not changed at a short time. Second, GraphSAGE uses an incremental scheme for representation learning, which means it only needs to consider the neighborhood of a new node during updating. GraphSAGE even does not need to figure out the high-order updating of nodes. In that case, GraphSAGE is quicker than DyGCN. However, GraphSAGE only considers how nodes interact with each other. It does not model the pattern of dynamic changes in graph. It gets worse performance than DyGCN in dynamic graph embedding task.

\noindent\textbf{Spectral DyGCN VS other methods.~}
Spectral DyGCN capture the dynamic propagation mechanism. It does not need to re-train any parameters during node updating, which cut down a lot of time consumption. Different from DyGCN, it updates the whole graph at each snap shot. Consequently, Spectral DyGCN has a more carefully consideration than DyGCN, however it consumes more time for updating than DyGCN.

%% file: 4_experiments.tex
\begin{table}[ht]
\centering\caption{Statistics of three evaluation datasets. (*-* means the number changes across different snapshot.)}
\begin{tabular}{|c|c|c|c|} 
\hline
Dataset & \#Nodes & \#Edges & \#Time steps \\
\hline
AS & 7716 & 8809-26467 & 60 \\
HEP-TH & 8557-14351 & 23318-47776 & 60 \\
Facebook &2085-2235 &9092-88642 & 36 \\
\hline
\end{tabular}\label{tab:dataset}
\end{table}

In this section, we detail our experiments, including the experiment setup, model comparison on effectiveness and efficiency, analysis on update order and visualizations.

\begin{table*}
\small
\centering\caption{Link prediction performance and running time (seconds) comparison of different methods. The best performance of dynamic embedding methods is highlighted. GCN is a static method which perform the upper bound in theory.}
\setlength{\tabcolsep}{2.4mm}{\begin{tabular}{l|ccc|ccc|ccc} 
\toprule
& \multicolumn{3}{c|}{AS} & \multicolumn{3}{c|}{HEP-TH} & \multicolumn{3}{c}{Facebook}\\
 & AUC & F1 & TIME & AUC & F1 & TIME & AUC & F1 & TIME\\
 \midrule

GraphSAGE & 0.5130 & 0.5421 & 0.2149 & 0.5576 & 0.5347 & 0.4134 & 0.5178 & 0.5151 & 0.0289 \\
TIMERS & 0.6763 & 0.6699 & 0.2235 & 0.5497 & 0.5326 & 0.2681 & 0.7228 & 0.5646 & 0.4337\\
DNE & 0.5818 & 0.6149 & 92.0739 & 0.6551 & 0.6185 & 111.3373 & 0.7395 & 0.6733 & 68.7416\\
 DynWalks & 0.8541 & 0.7674 & 13.5219 & 0.8867 & 0.8229 & 17.5355 & 0.7488 & 0.7231 & 8.1941\\
DyGCN & 0.8621 & 0.7694 & 0.3138 & 0.8964 & 0.8679 & 0.5973 & 0.7544 & 0.7430 & 0.0350\\
Spectral DyGCN &\textbf{0.8738} & \textbf{0.7710} & 0.4374 & \textbf{0.9039} & \textbf{0.8806} & 0.7549 & \textbf{0.7697} & \textbf{0.7482} & 0.0548\\
\hline \hline
GCN & 0.8935 & 0.7731 & 221.8825 & 0.9324 & 0.8983 & 453.9417 & 0.8184 & 0.7493 & 143.9285\\
\bottomrule
\end{tabular}}
\label{tab:lp}
\end{table*}

\begin{figure}[ht]
\centering
\includegraphics[width=0.6\linewidth]{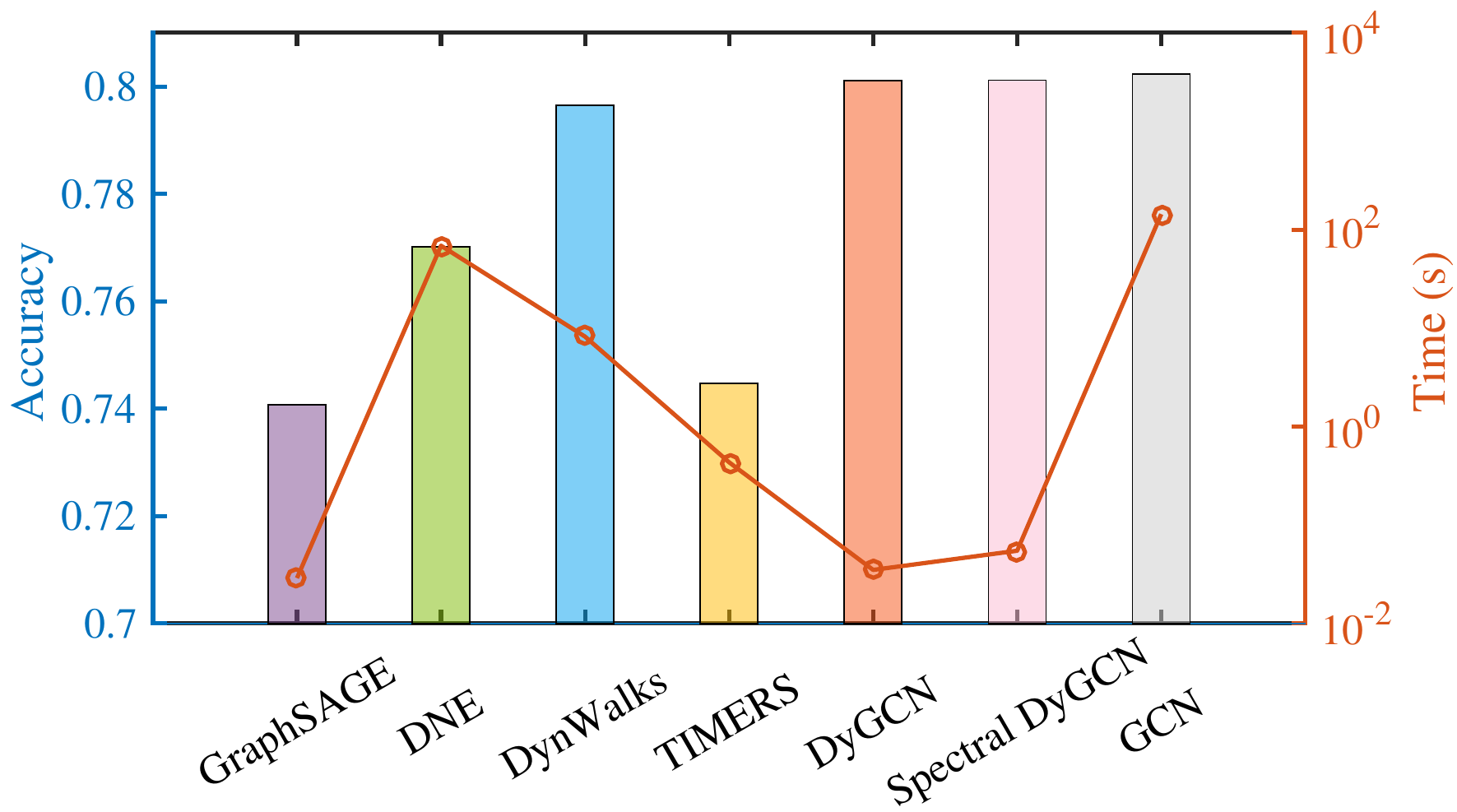}
\caption{Performance comparison of node classification task on Facebook dataset along with running time (seconds).}
\label{fig:nc}
\end{figure}

\subsection{Experiment Setup}
\noindent\textbf{Datasets.~}
Following \cite{goyal2018dyngem,du2018dynamic}, we conduct experiments on the following three real-world datasets, whose statistics can be found in Table \ref{tab:dataset}.
\begin{itemize}
 
  \item \textbf{Autonomous Systems (AS)} \cite{leskovec2014snap} is a communication network from the BGP (Border Gateway Protocol) logs. 
The dataset contains instances spanning from 1997 to 2000. 
For our evaluation, we select 60 consecutive snapshots in the dataset.
  \item \textbf{HEP-TH} \cite{gehrke2003overview} is a dataset containing abstracts of papers in High Energy Physics Theory conference from 1993 to 2003. We create a collaboration network for papers published in each month. Likewise, we select 60 consecutive snapshots in the dataset.
  \item \textbf{Facebook} \cite{traud2012social} is a social network dataset on Facebook website. We select a subset of Amherst College. To generate a sequence of dynamic networks, we randomly sample a sub-graph from the original network as the initial networks. Each time step $t$, we add a fixed size of new nodes and edges.
\end{itemize}

\noindent\textbf{Compared Methods.~}
We compare our proposed model DyGCN with the following methods:
\begin{itemize}
\item \textbf{GraphSAGE}~\cite{hamilton2017inductive} is an inductive GCN-based method which could deal with new nodes, but it does not model the changes between time steps.
\item \textbf{TIMERS}~\cite{zhang2018timers} is a dynamic embedding model based on incremental Singular Value Decomposition (SVD) with theoretically instructed maximum-error-bounded restart.
\item \textbf{DNE}~\cite{du2018dynamic} is an extension of skip-gram based methods to dynamic setting with high efficiency.
\item \textbf{DynWalks}~\cite{hou2019dynwalks} is an extension of Deepwalk~\cite{perozzi2014deepwalk}, which can dynamically and efficiently learn embeddings based on the selected ones with high efficiency. 
\item \textbf{GCN}~\cite{kipf2016variational,kipf2016semi} is a prevalent static embedding model and also the static counterpart of our proposed model DyGCN. In dynamic scenarios, we treat each snapshot of the dynamic graphs as a static graph and utilize GCN to learn node embeddings on it. The node degrees are used as node features.
\end{itemize}

It is important to note that we use GCN as the static base model to obtain the original embeddings at each time step in DyGCN, and then learn the embeddings at next time steps. 
The dimension of node embeddings $d$ is 100 for all methods.
The maximum order of update $K$ is set as 2 by default.

\noindent\textbf{Tasks and Evaluation Metrics.~}
We conduct experiments and compare the methods on the following two tasks:
\begin{itemize}
\item \textbf{Link Prediction}: It aims to test how well the learned embeddings can predict unobserved edges.
We measure the edge proximity based on the cosine distance between embeddings of pairs of nodes.
To achieve this, we evaluate the ability of predicting future links at $t$+1 time step using embeddings we learn at $t$ time step. 
We select \textbf{AUC} (Area Under the ROC curve) and \textbf{F1-score} as the evaluation metrics for the this task. As discussed in Section \ref{sec:related_work_dge}, DyGCN is not designed for link prediction. However, link prediction task is still a traditional task to evaluate the performance of node embedding. 
\item \textbf{Node Classification}: In order to test the classification ability of the embeddings, we use Logistic Regression as the classifier with the input of node embeddings. Here, the we use the $Z^{0}$ at the initial time step as the training data for the parameters of Logistic Regression, and hold the parameters unchanged at the followint time step.   
\textbf{Accuracy} is the evaluation metric of node classification. We summarize the predicted result of all the nodes at each times by using the updating embeddings. The hyper-parameters of Logistic Regression is the default setting in scikit-learn\footnote{https://scikit-learn.org/stable/modules/generated/sklearn.linear\_model.LogisticRegression.html}.
\end{itemize}

For the three datasets, we use the first half of time steps to constitute the training set, and the second half of time steps are used as test set.
The reported results are averaged over all the snapshots in test set.

\begin{figure*}[h]
\centering
\subfigure[AS]{
\begin{minipage}[b]{0.32\textwidth}
\label{fig:longterm_as} 
\includegraphics[width=1\textwidth]{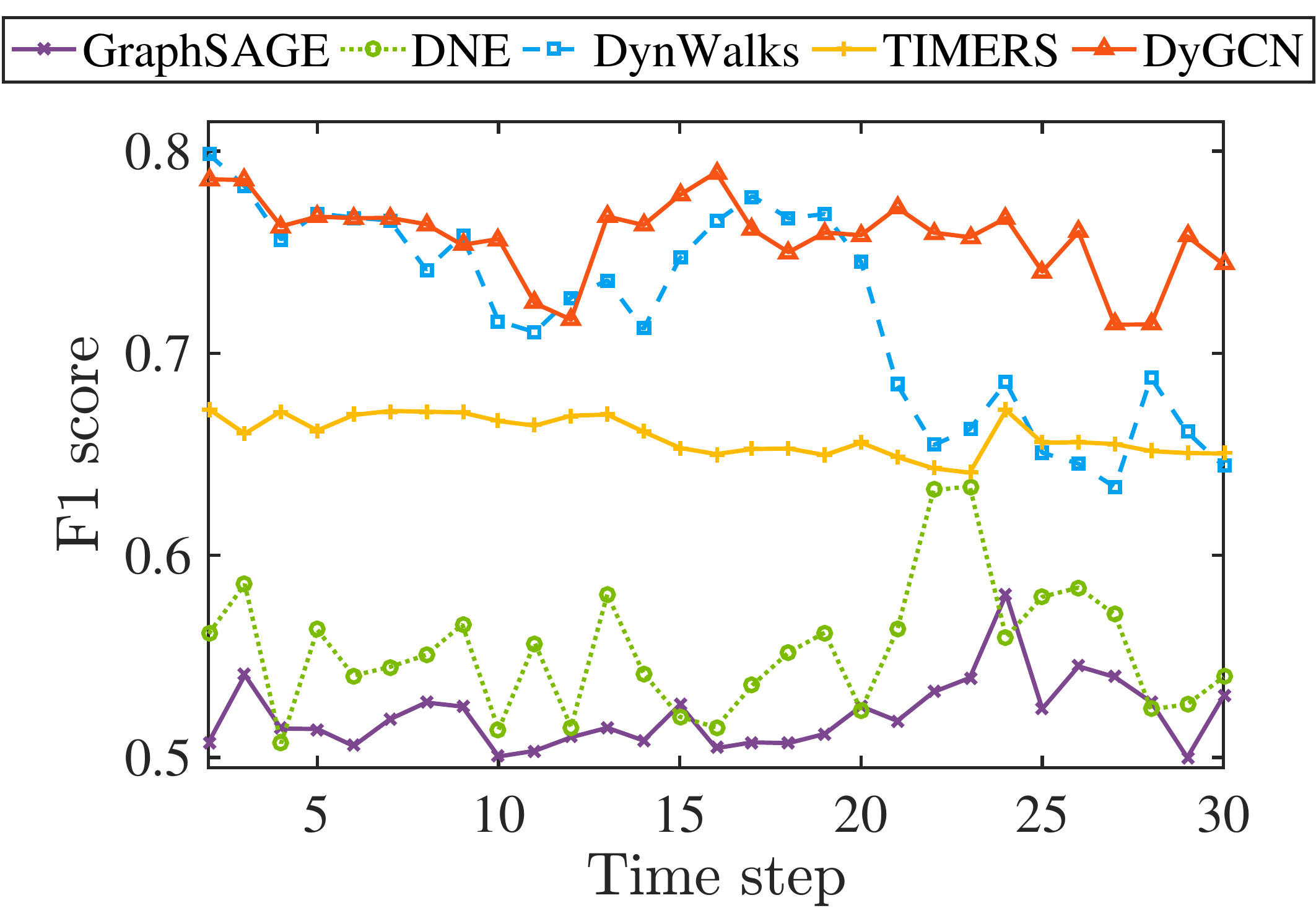}
\end{minipage}%
}%
\subfigure[HEP-TH]{
\begin{minipage}[b]{0.32\textwidth}
\label{fig:longterm_hep} 
\includegraphics[width=1\textwidth]{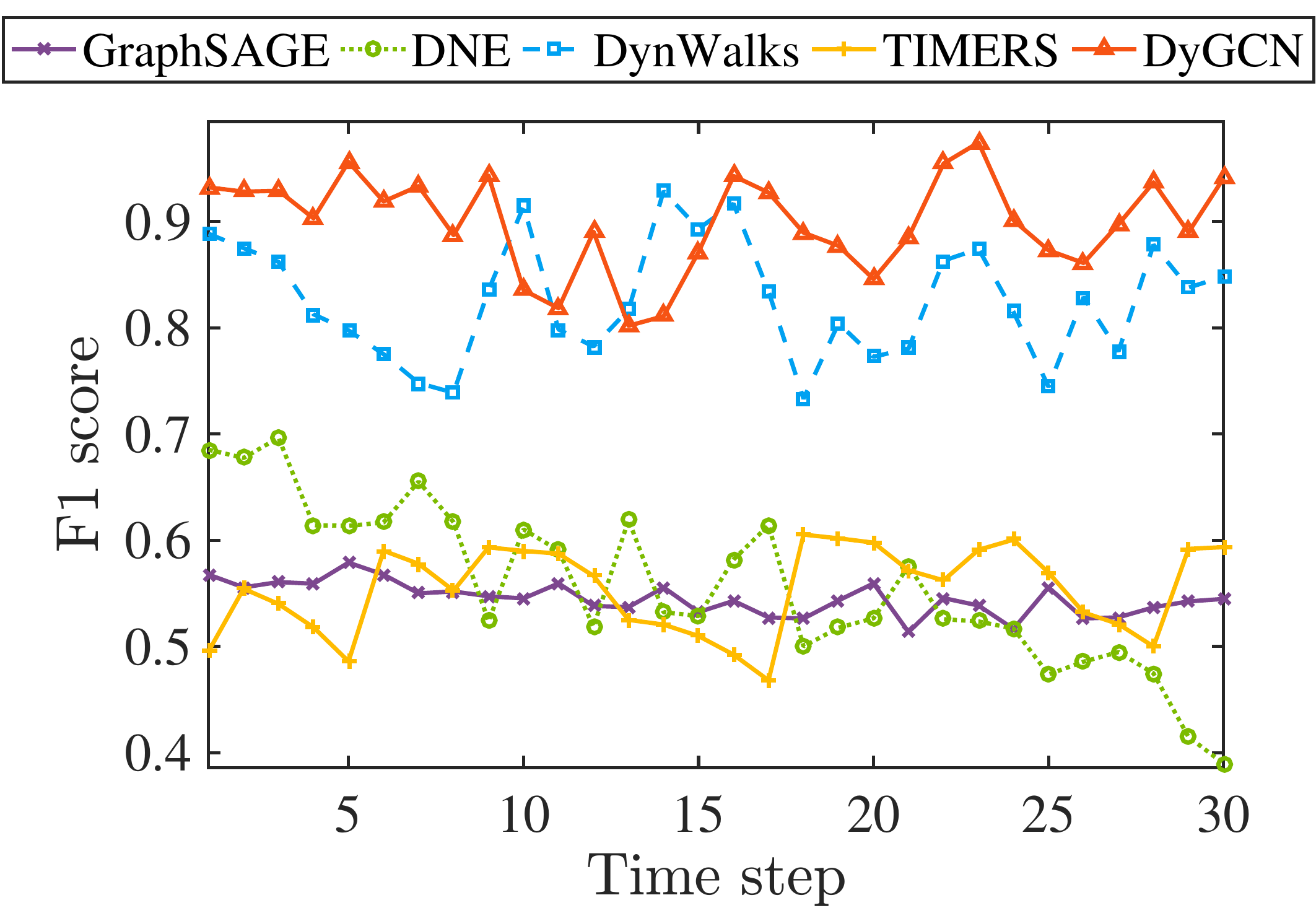}
\end{minipage}%
}%
\subfigure[Facebook]{
\begin{minipage}[b]{0.32\textwidth}
\label{fig:longterm_amherst} 
\includegraphics[width=1\textwidth]{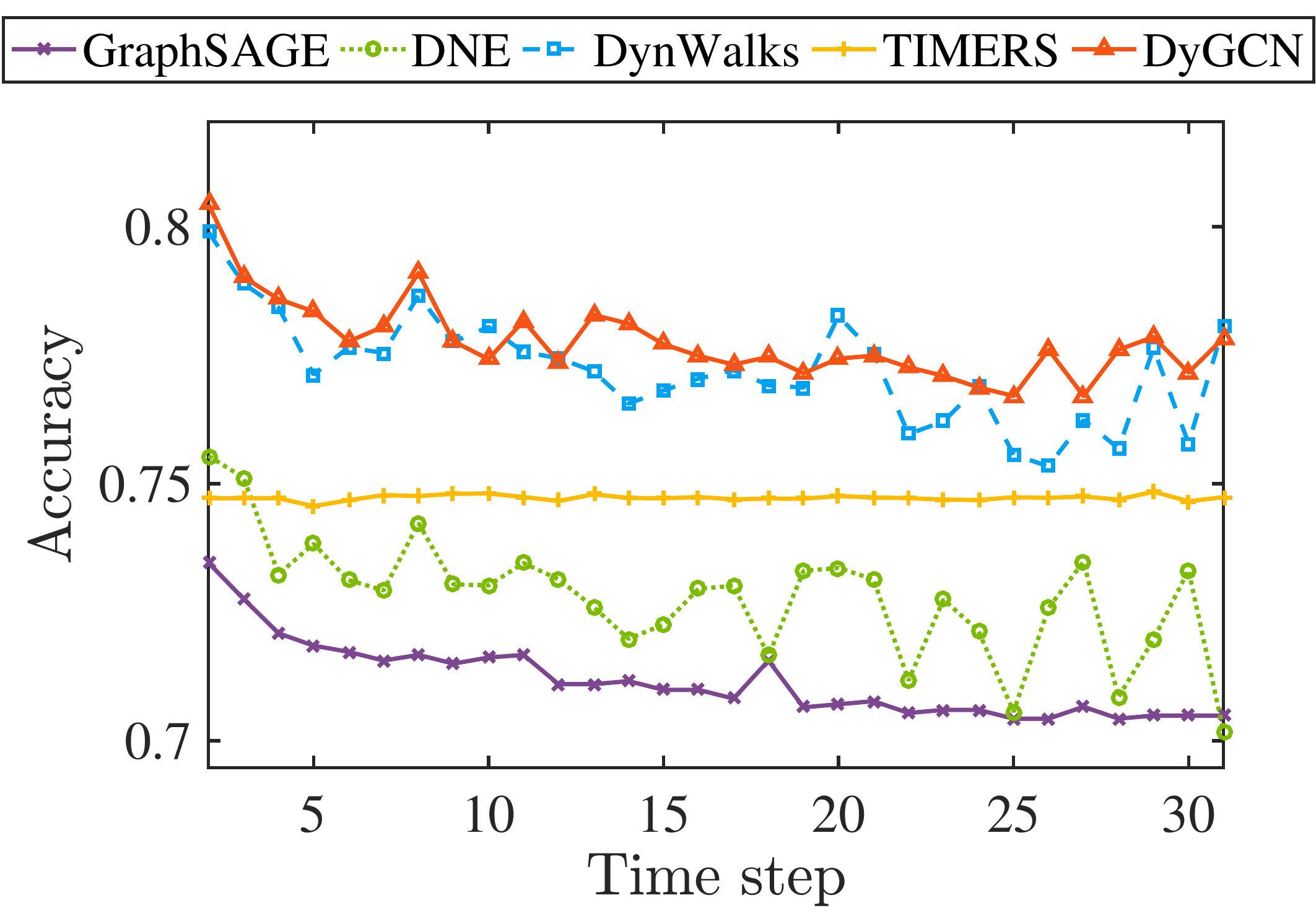}
\end{minipage}%
}%
\caption{Performance comparison of long term link prediction task w.r.t. F1-score on AS, HEP-TH dataset and node classification task on Facebook dataset.}
\label{fig:long-term}
\end{figure*}

\subsection{Results and Analysis}
Table~\ref{tab:lp} shows the performance of link prediction task and also running time (seconds) of evaluation at each time step on three datasets, while Figure~\ref{fig:nc} illustrates the performance of node classification task on Facebook dataset.
Note that GCN which achieves the best performance and consumes the most running time is the base model of DyGCN.
Its performance is the upper bound of DyGCN in theory.

Jointly analyzing Table~\ref{tab:lp} and Figure~\ref{fig:nc}, we find that GraphSAGE achieves the worst performance on all cases. 
This is expected since it is originally designed for static embedding and fails to capture the changing patterns during time steps. However, GraphSAGE is the fastest method among all, its incremental computation ensures its efficiency, i.e., each new node only needs to aggregate its neighbors' information for representation.  
TIMERS achieves a modest performance with a little running time. It only updates a part of nodes at each time step, and it does not need any optimization progress during updating. Its performance is modest because TIMERS is based on matrix perturbation theory which only works when the adjacency matrix does not change a lot at every time step \cite{zhang2018timers}.  
Both DNE and DynWalks need SGD for updating nodes, so they consume much more time than other methods. 
DNE achieves better performance than TIMERS on HEP-TH and Facebook dataset, at the cost of much more time consumption. On AS dataset, DNE is not obviously better than TIMERS. From our perspective, the graph in AS is much sparser than the other datasets and an adding edge will not affect the adjacency matrix much, so that TIMERS gets comparatively better results on AS.
DyWalks performs better than DNE. DyWalks is mainly motivated by incremental skip-gram model which is well-designed for efficiency and effectiveness. It only update the nodes who may be reached by random walk with the start of 1st-order nodes. 
Our proposed DyGCN consistently yields the best performance on all cases with a limited time cost, which shows its superiority in terms of accuracy and efficiency. It gets the comparative performance as DynWalks, and the same level of running time as GraphSAGE and TIMERS.
Satisfactorily, compared with the base model GCN, DyGCN can achieve competitive performance on both link prediction task and node classification task with much less running time.
Quantitatively, DyGCN is 692$\times$, 759$\times$ and 4110$\times$ faster than its static counterpart GCN on AS, HEP-TH and Facebook dataset, respectively.
This is reasonable since the number of updated nodes is much less than the number of all the nodes, besides there is no need of training process during updating.
Spectral DyGCN performs better than DyGCN and it consumes acceptable time compared with other methods. It makes sense since Spectral DyGCN update the whole graph instead of only a part of nodes of DyGCN. 
Note again that GCN which achieves best performance and most running time is the base model of DyGCN.
Its performance is the upper bound of DyGCN in theory.
To summarize, this further verifies the effectiveness and efficiency of our proposed DyGCN, since our proposed models achieve the nearest results with GCN and use the much less time consumption.


\begin{figure*}[t]
\subfigure[AS]{
\begin{minipage}[b]{0.32\textwidth}
\label{fig:as_korder} 
{\includegraphics[width=1\textwidth]{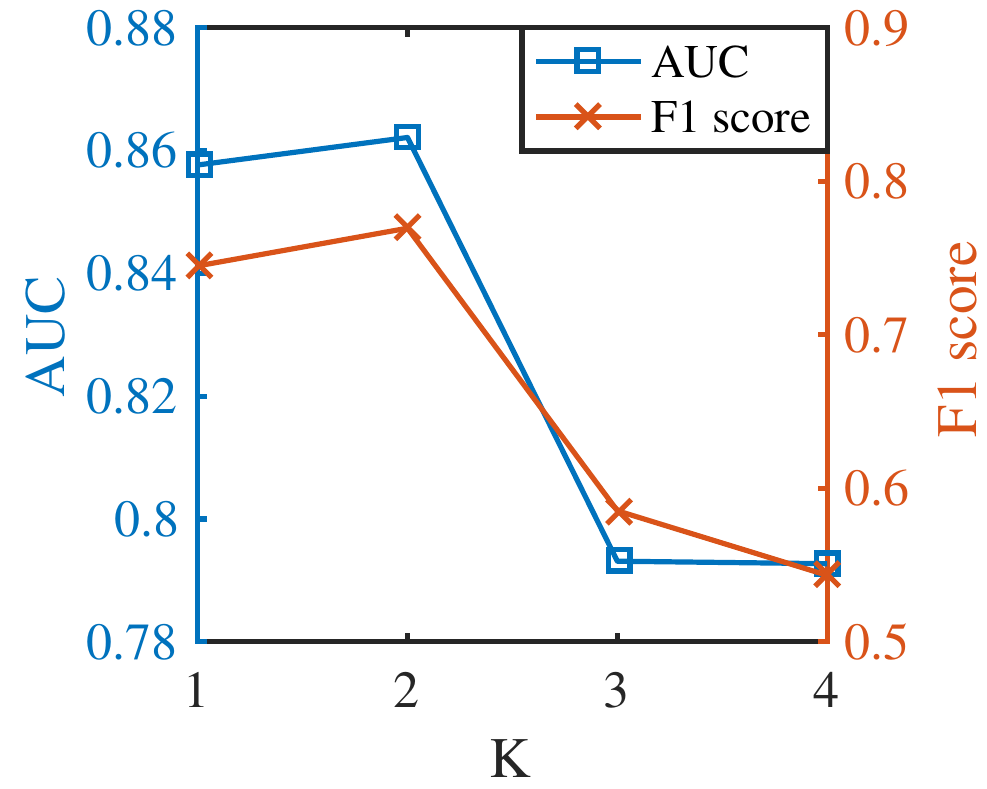}

\includegraphics[width=0.93\textwidth]{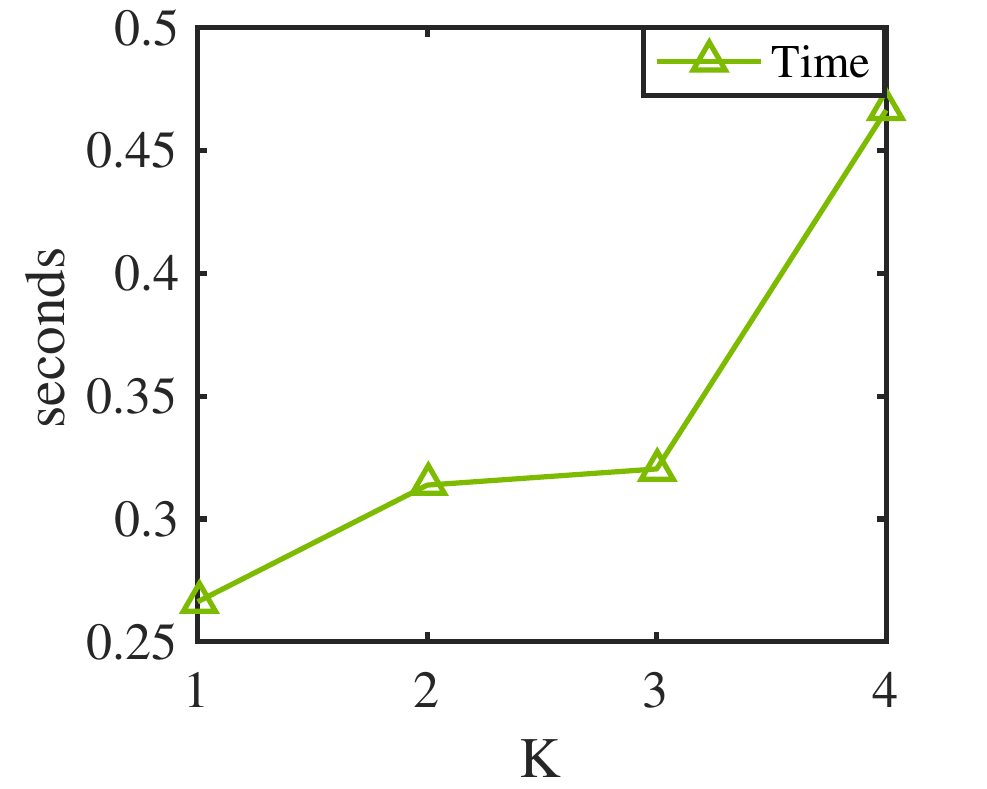}}
\end{minipage}%
}%
\subfigure[HEP-TH]{
\begin{minipage}[b]{0.32\textwidth}
\label{fig:hep_korder} 
\includegraphics[width=1\textwidth]{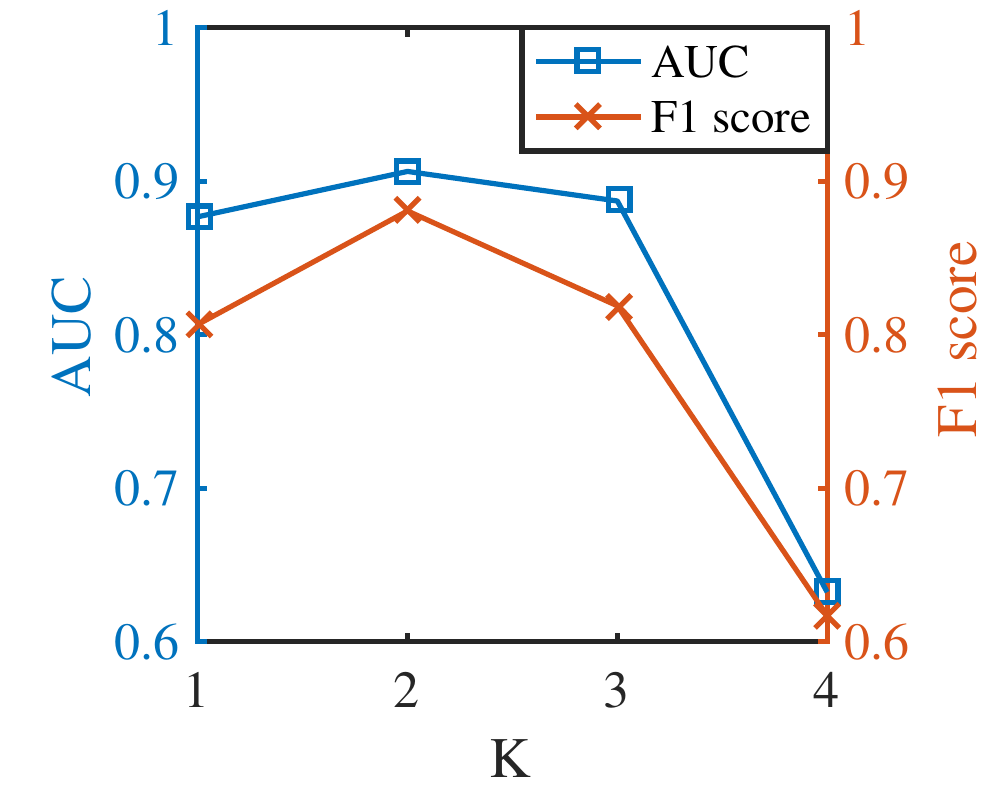}

\includegraphics[width=0.93\textwidth]{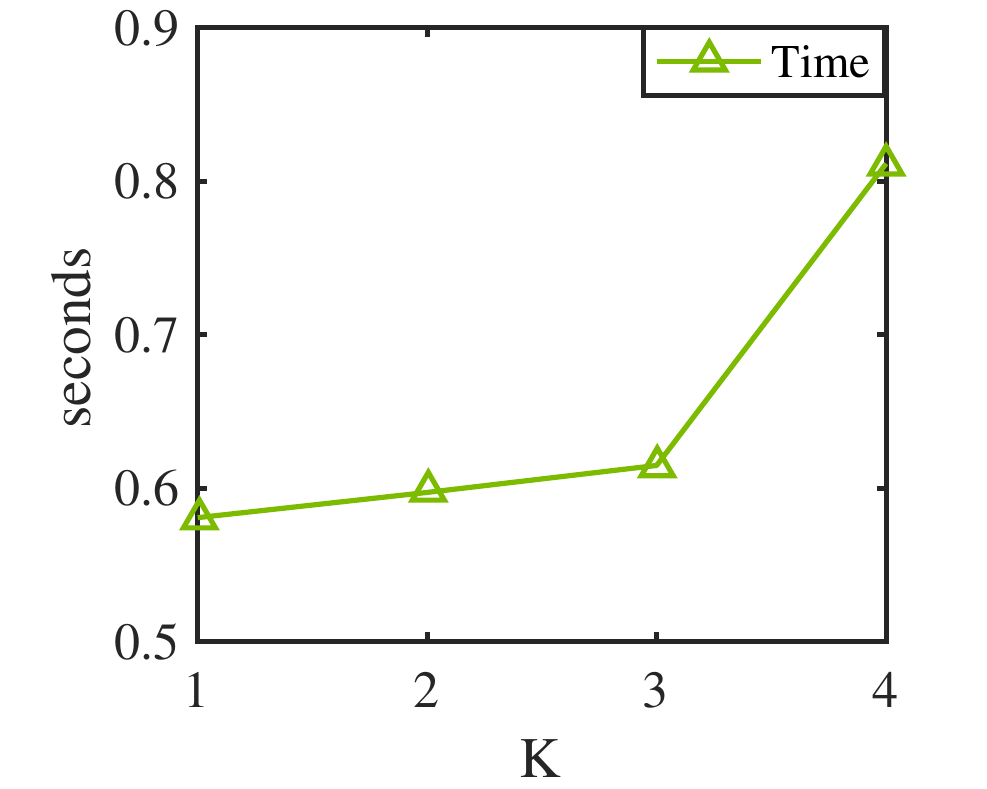}
\end{minipage}%
}%
\subfigure[Facebook]{
\begin{minipage}[b]{0.32\textwidth}
\label{fig:facebook_korder} 
{\includegraphics[width=1\textwidth]{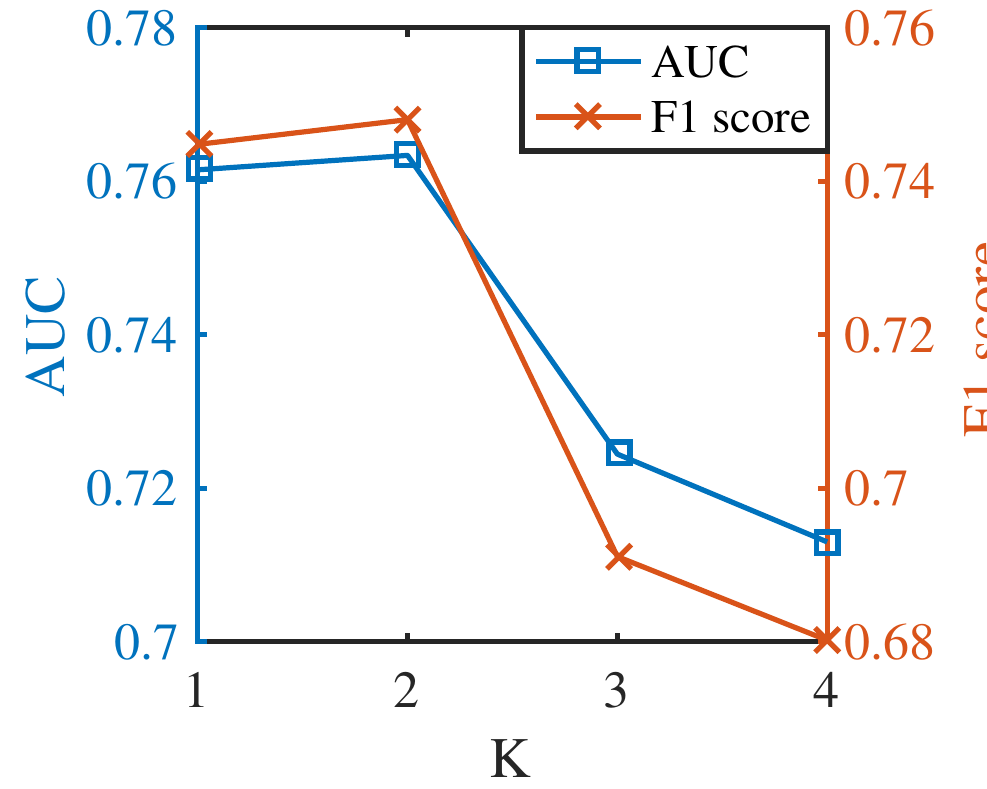}

\includegraphics[width=0.93\textwidth]{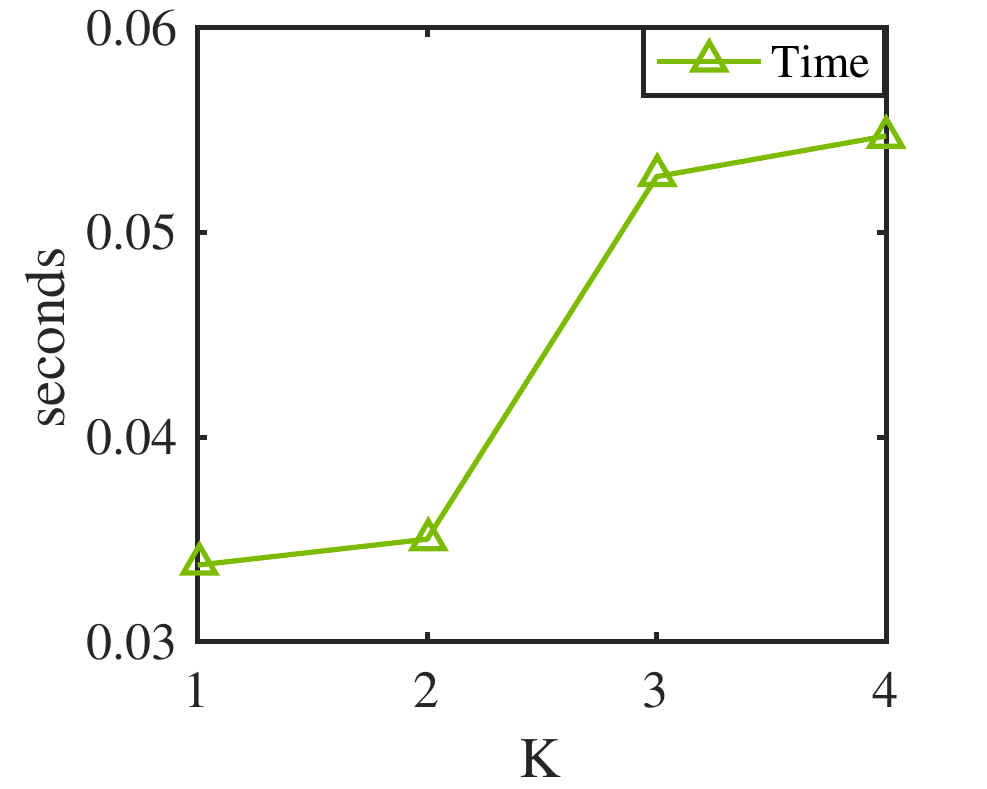}}
\end{minipage}%
}%
\caption{Performance and running time w.r.t. different orders of update $K$.}
\label{fig:korder}
\end{figure*}


\subsection{Long-term Dynamic Embedding}
To test the robustness of our proposed model in case of cumulative errors in long-term embedding, we compare these models on long-term tasks of both link prediction and node classification.
The long-term tasks are basically same as the link prediction and node classification tasks described above, except that only the initial node embeddings at $t=0$ are given. We simply use the calculated node embeddings at $t-1$ as the initial node embeddings at $t$. 
At the consecutive time steps, we use the node embeddings at previous time step learned by DyGCN as the given node embeddings to calculate the node embeddings at next time step.
In specific, we aim to learn the node embeddings at each time step $t$ varying from $1$ to $30$, only given the node embedding at $t=0$.
The results at every step are shown in Figure \ref{fig:long-term}.
As can be seen, GraphSAGE gets the comparatively worst performance among all, since it does not gain any knowledge from dynamic changes. TIMERS achieves a high accuracy result in AS, which has been discussed above that TIMERS may be exactly suitable for a sparser graph like AS dataset. TIMERS, GraphSAGE and DyGCN who do not need optimization progress are stabler across all time steps than other methods. It is may because of the unstable randomness coming from  optimization progress. It is obvious that all the methods drop along with the time steps. 
However, it is interesting to discover that the performance tends to have an unpredictable  vibration during $20 \sim 30$ time step both in AS and Facebook datasets. While there is not too big vibration in HEP-TH. In our opinion,  the sparser a graph is, the harder it is to represent each node. 
Our DyGCN and DynWalks consistently outperform all other methods on all datasets. 
From the holistic view, DyGCN performs better than DynWalks, especially when it is long time away from the initialization. 
Moreover, we can observe that the performance of DynWalks drops sharply on AS dataset, while DyGCN is much more stable.  
Overall, we can draw the conclusion that DyGCN is robust on long-term tasks, which is very important in real world applications.

\subsection{Update Order}
To investigate how the maximum updated order $K$ affects the model accuracy and efficiency, we vary it in the range of $\left \{1,2,3,4\right \}$.
Figure~\ref{fig:korder} summarizes the experimental results. The results of DyGCN are performed on the left side, and the time consumption are on the right side.
We can observe that with the increasing of update order, the performance first increases and then drops sharply after $K>2$ on the dataset of AS and Facebook. 
This suggests that the change affects the 1st-order and 2nd-order influenced nodes most and the influence on further nodes is relatively ignorable.
Propagating the change to the further nodes and updating them is superfluous and can deteriorate the performance.
However, both AUC and F1-score are maintain relatively high until $K=3$ in HEP-TH whose graph is much more dense and the edges changes much fewer than other datasets. As can be seen in Table \ref{tab:average_num_between_hops}, HEP-TH has extremely less 1st-order nodes, and much more 2nd,3rd-order (especially 3rd-order) nodes than other datasets. It inspires us that when we use DyGCN in a graph which gets less 1st-order nodes and more 3rd-order nodes, we many need more orders of update. It will be discussed in the following section.
In addition, the increasing of $K$ brings about more time consumption especially after $K>3$.
This is reasonable since the number of influenced nodes increase exponentially along with the increase of update order. 
When $K >3$, the time consumption even higher than Spectral DyGCN.  
Therefore, we only need update 1st-order and 2nd-order influenced nodes, which is the best in terms of both effectiveness and efficiency.

\begin{figure*}[ht]
\centering
\subfigure[AS]{
\begin{minipage}[b]{0.34\textwidth}
\label{fig:hidden_as} 
\includegraphics[width=1\textwidth]{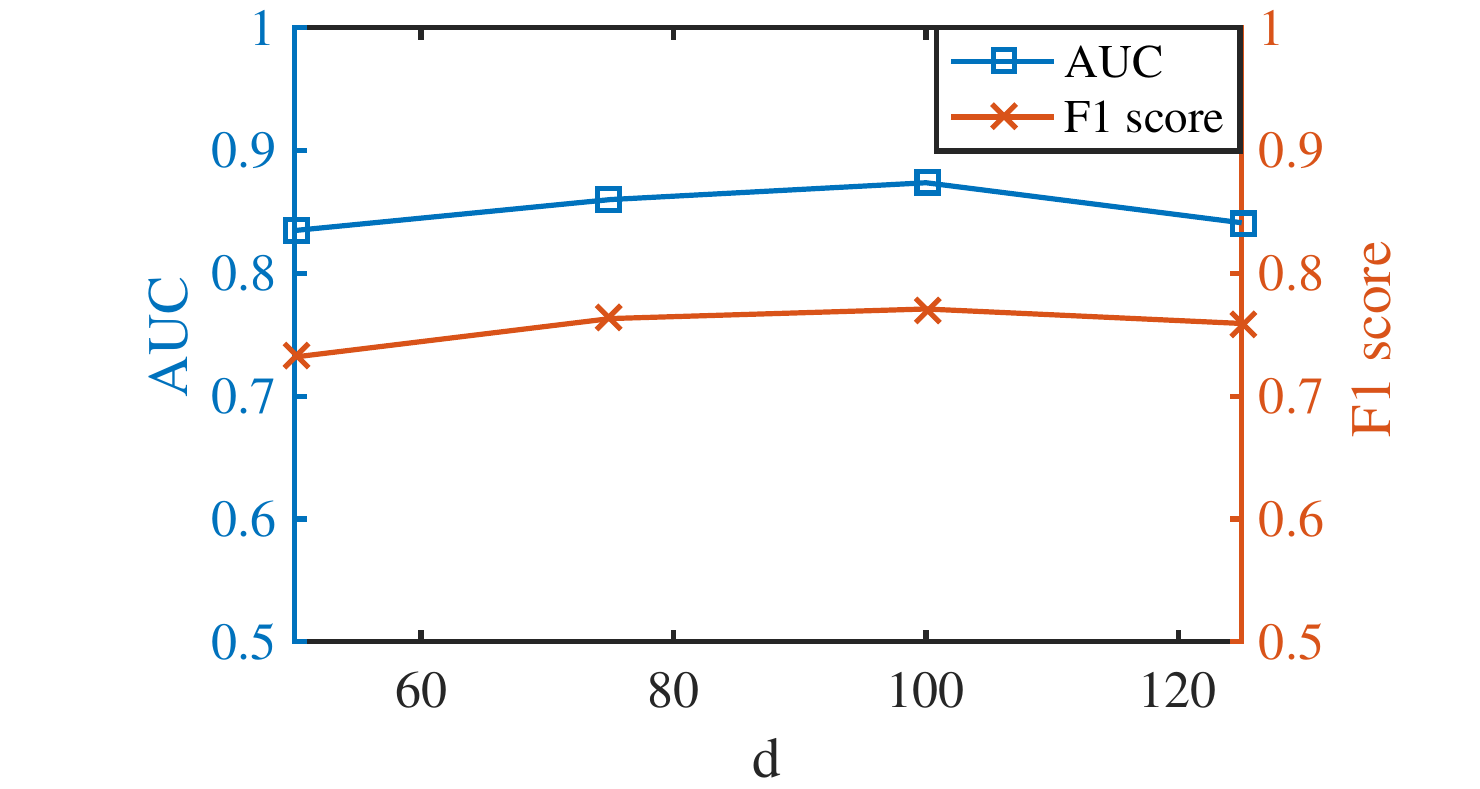}
\end{minipage}%
}%
\subfigure[HEP-TH]{
\begin{minipage}[b]{0.34\textwidth}
\label{fig:hidden_hep} 
\includegraphics[width=1\textwidth]{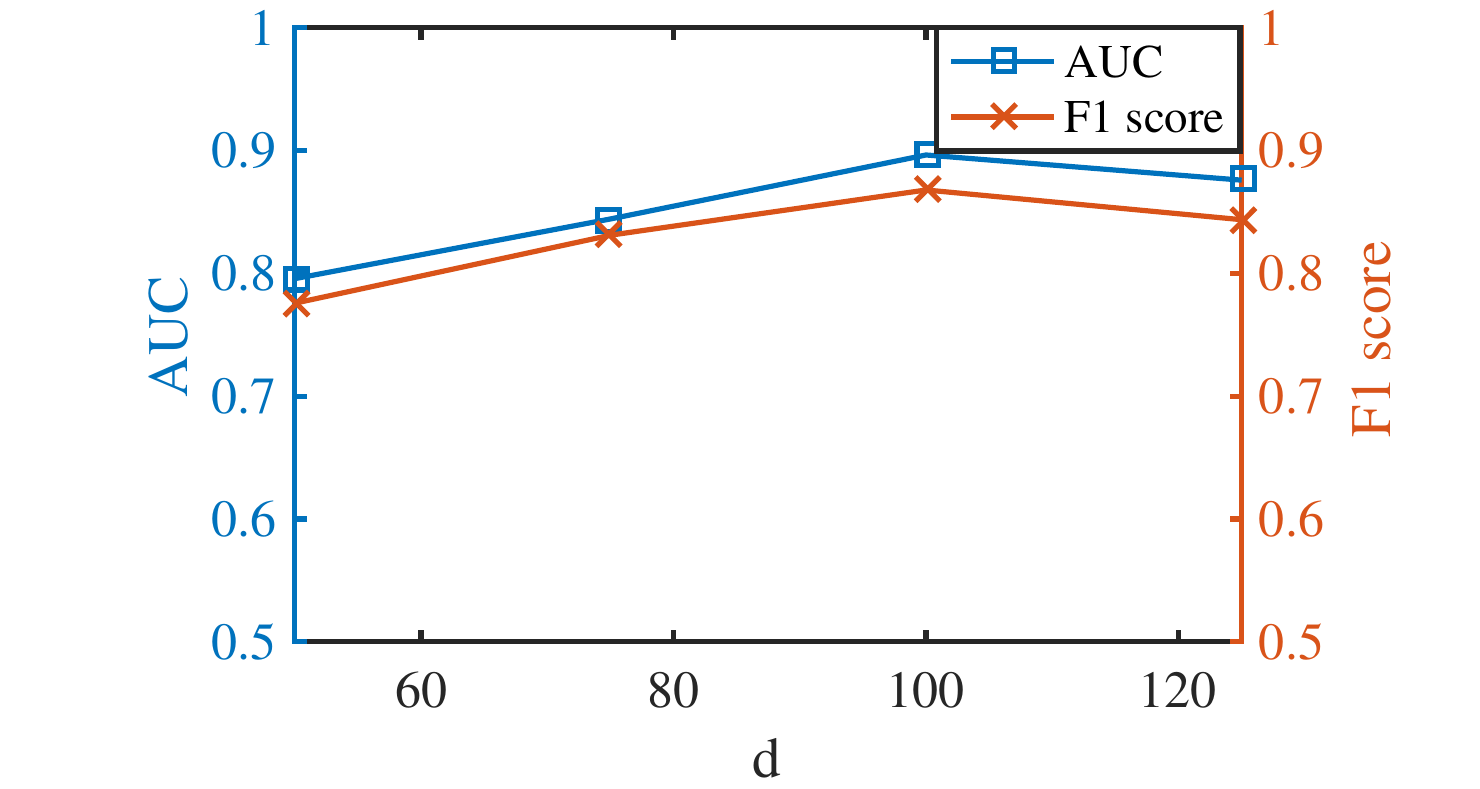}
\end{minipage}%
}%
\subfigure[Facebook]{
\begin{minipage}[b]{0.34\textwidth}
\label{fig:hidden_amherst} 
\includegraphics[width=1\textwidth]{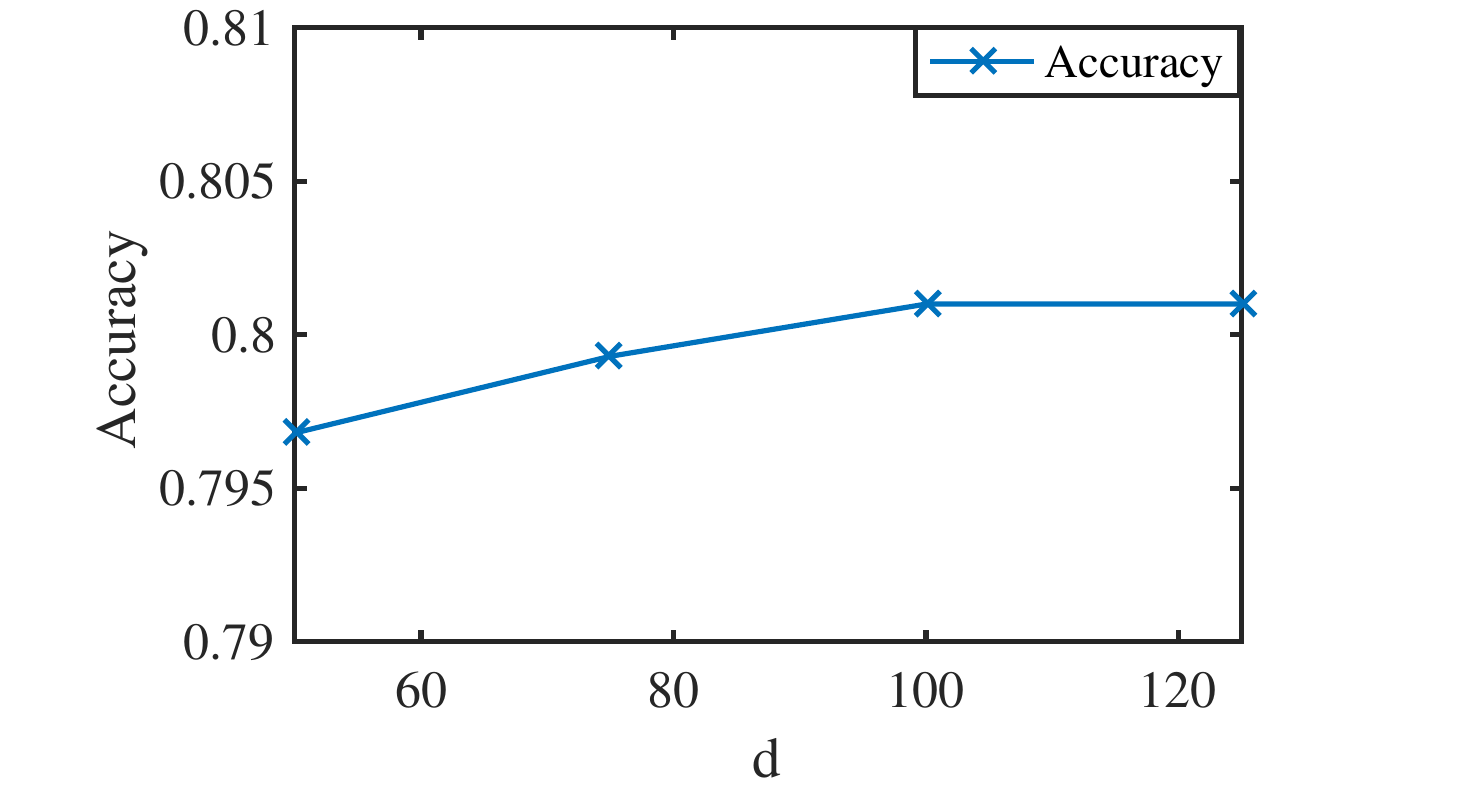}
\end{minipage}%
}%
\caption{The performance of DyGCN of different hidden vector size $d$ on three datasets.}
\label{fig:hidden_vector}
\end{figure*}

\subsection{Dimensionality of Hidden Vectors}
We try our models with different embedding dimensionalities $d$ of latent vectors. We do experiments on different hidden vector sizes as 50, 75, 100 and 125. 
As illustrated in Figure \ref{fig:hidden_vector}, DyGCN performs relatively stable with varying dimensionalities among all the datasets. When the embedding dimensionality is high, DyGCN tends to be over-fitting. 
DyGCN get the best performance  with $d=100$  in all of the three datasets.
We compare the datasets of AS and HEP-TH, and find that HEP-TH may needs higher-dimensional vectors to represent than AS for the reason that AS has a sharper decrease from 100 to 120 than HEP-TH.  
It may be because that HEP-TH has more nodes that AS which makes it more difficult to distinguish different nodes. 
 

\begin{figure*}[t]
\subfigure[AS]{
\begin{minipage}[]{1\textwidth}
\label{fig:as_vishops} 
\centerline{\includegraphics[width=0.25\textwidth]{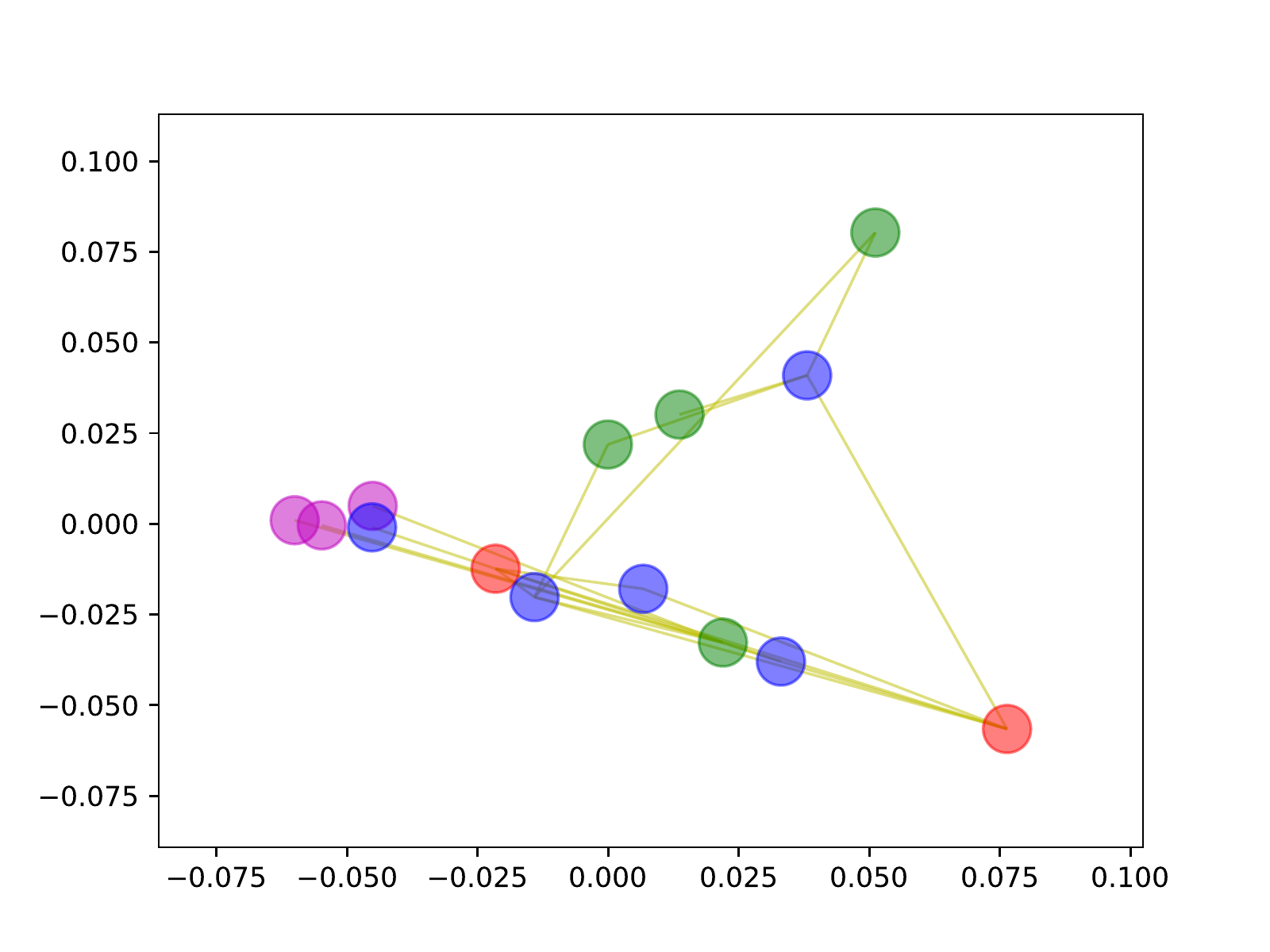}\includegraphics[width=0.25\textwidth]{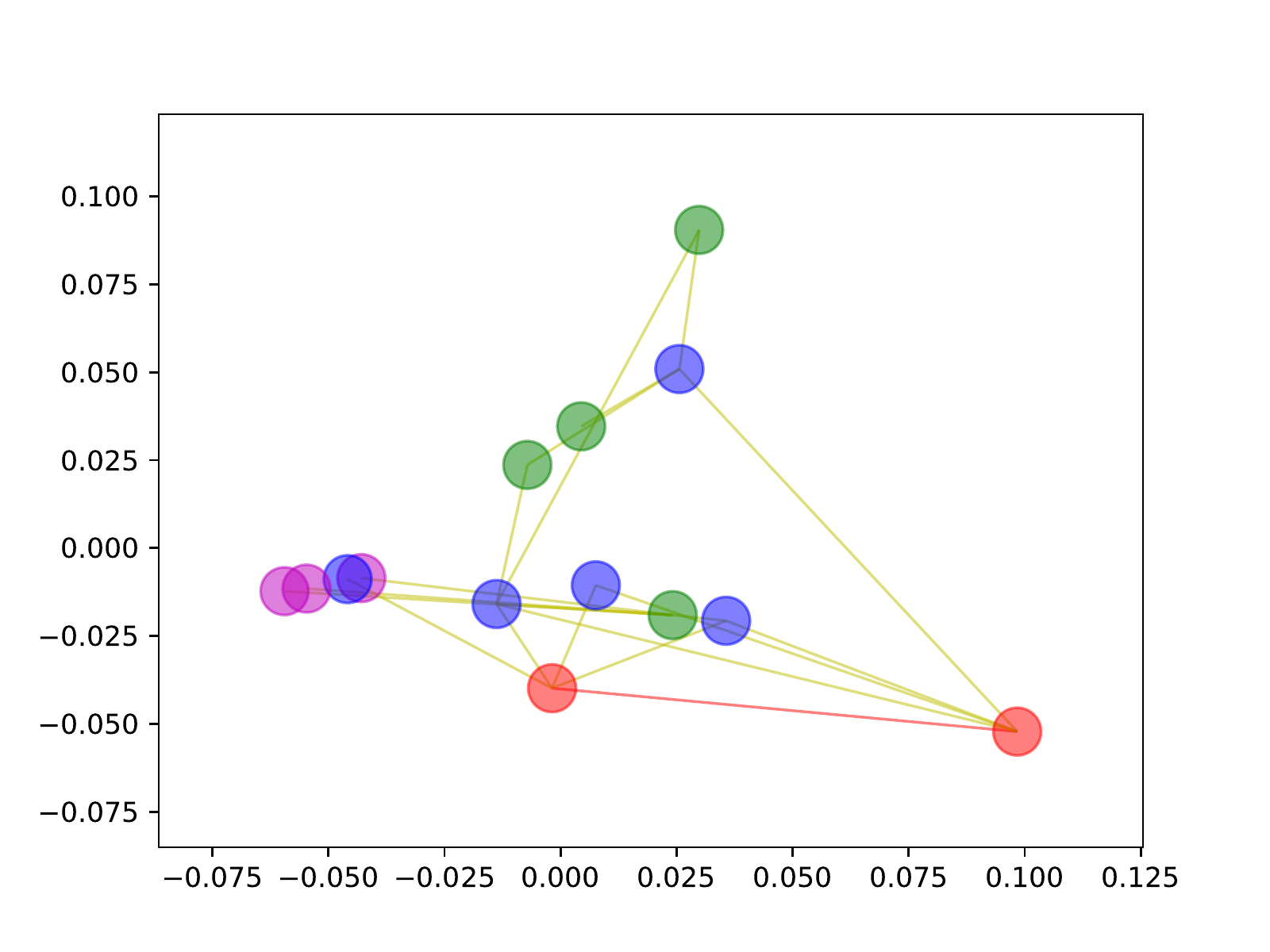}\includegraphics[width=0.25\textwidth]{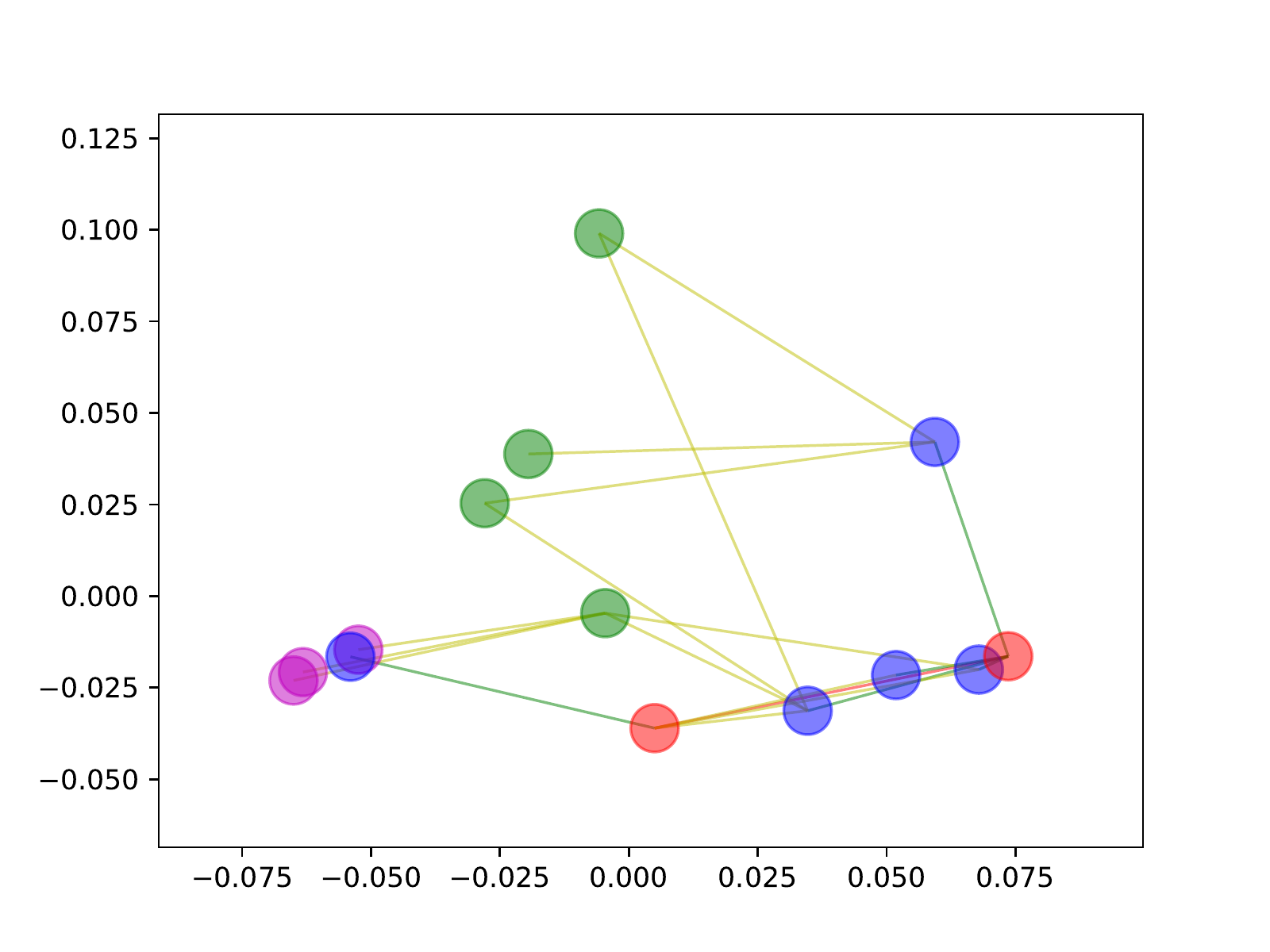}\includegraphics[width=0.25\textwidth]{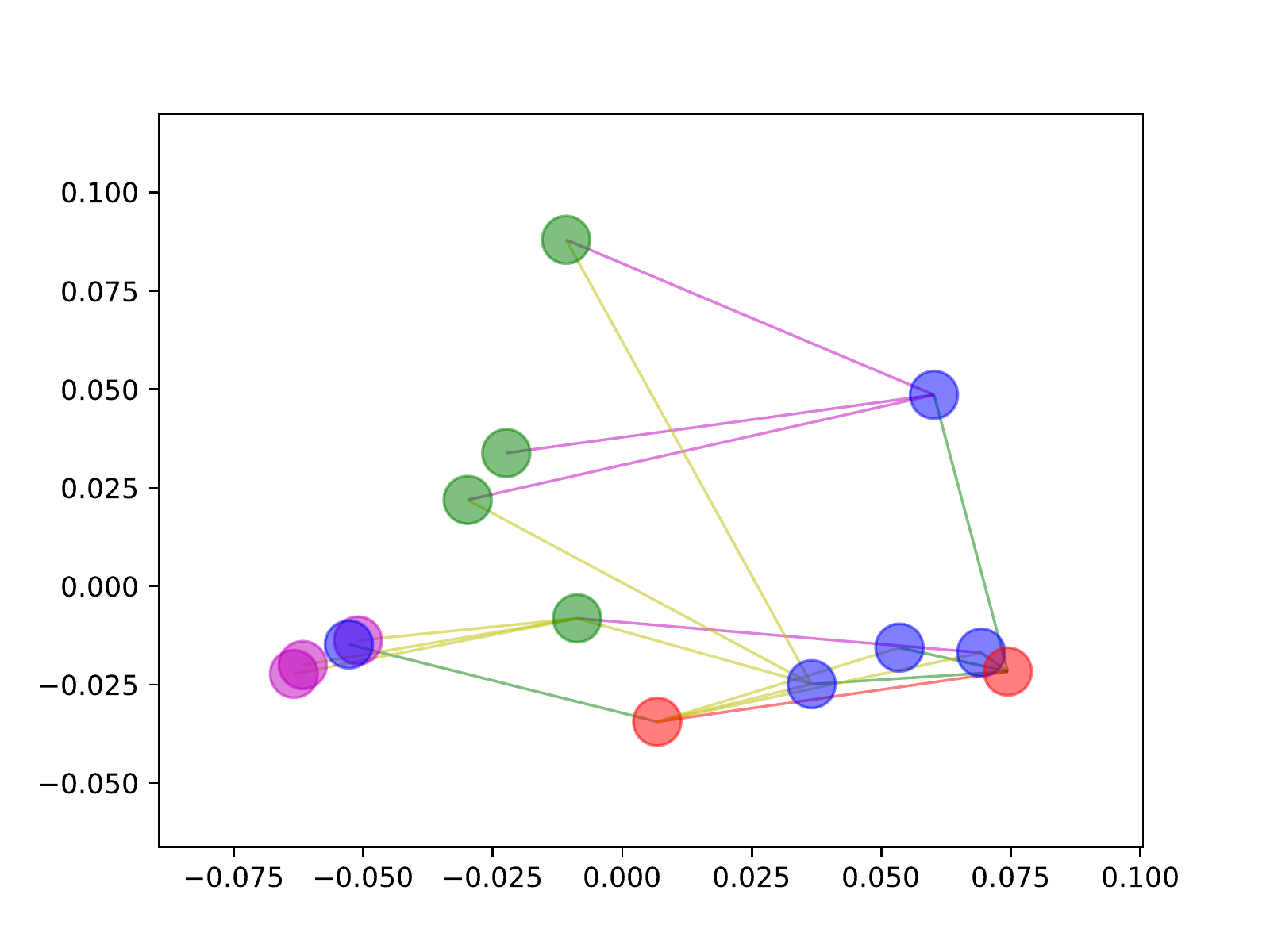}}
\end{minipage}%
}%

\subfigure[HEP-TH]{
\begin{minipage}[b]{1\textwidth}
\label{fig:hep_vishops} 
\centerline{\includegraphics[width=0.25\textwidth]{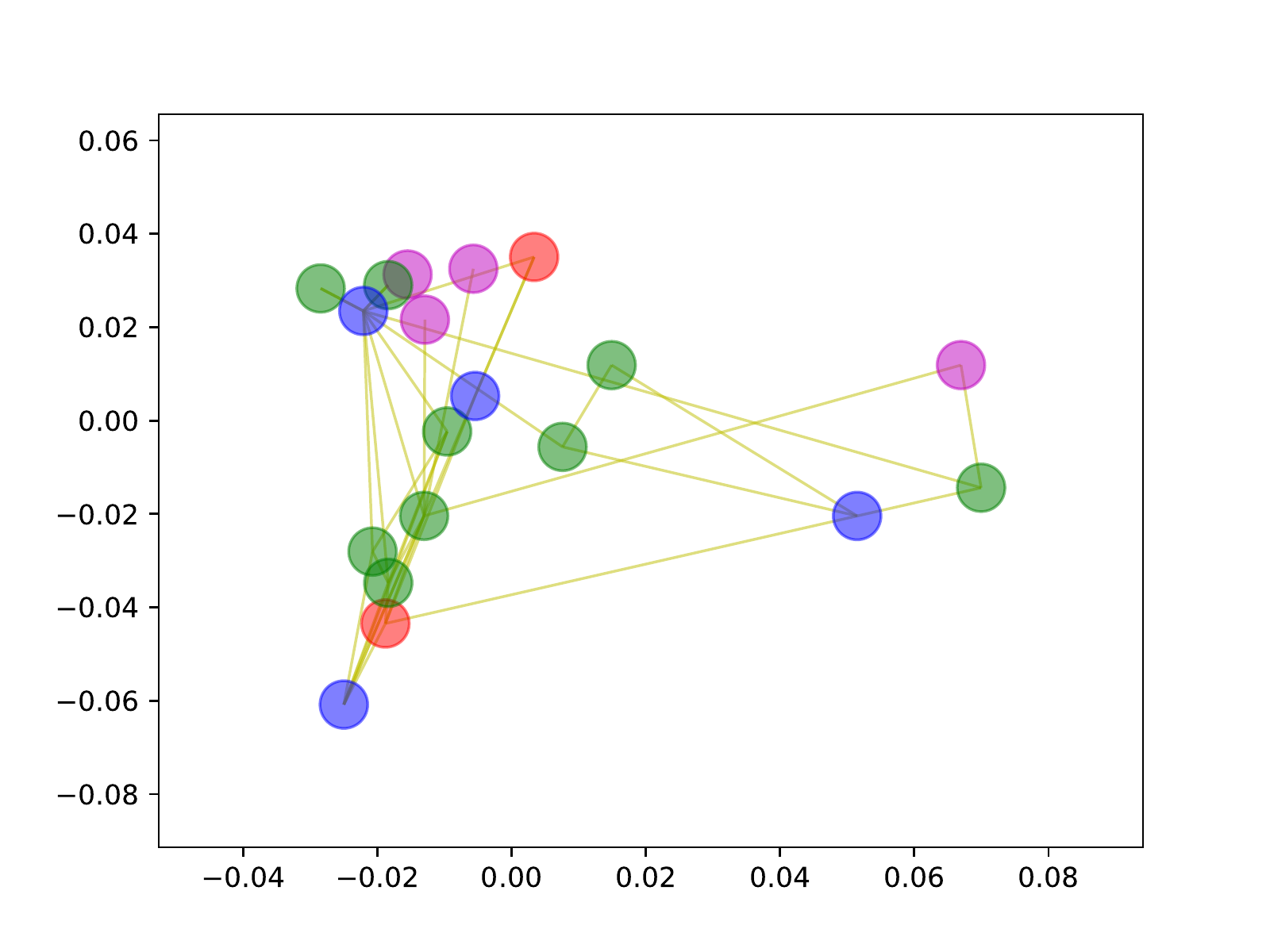}\includegraphics[width=0.25\textwidth]{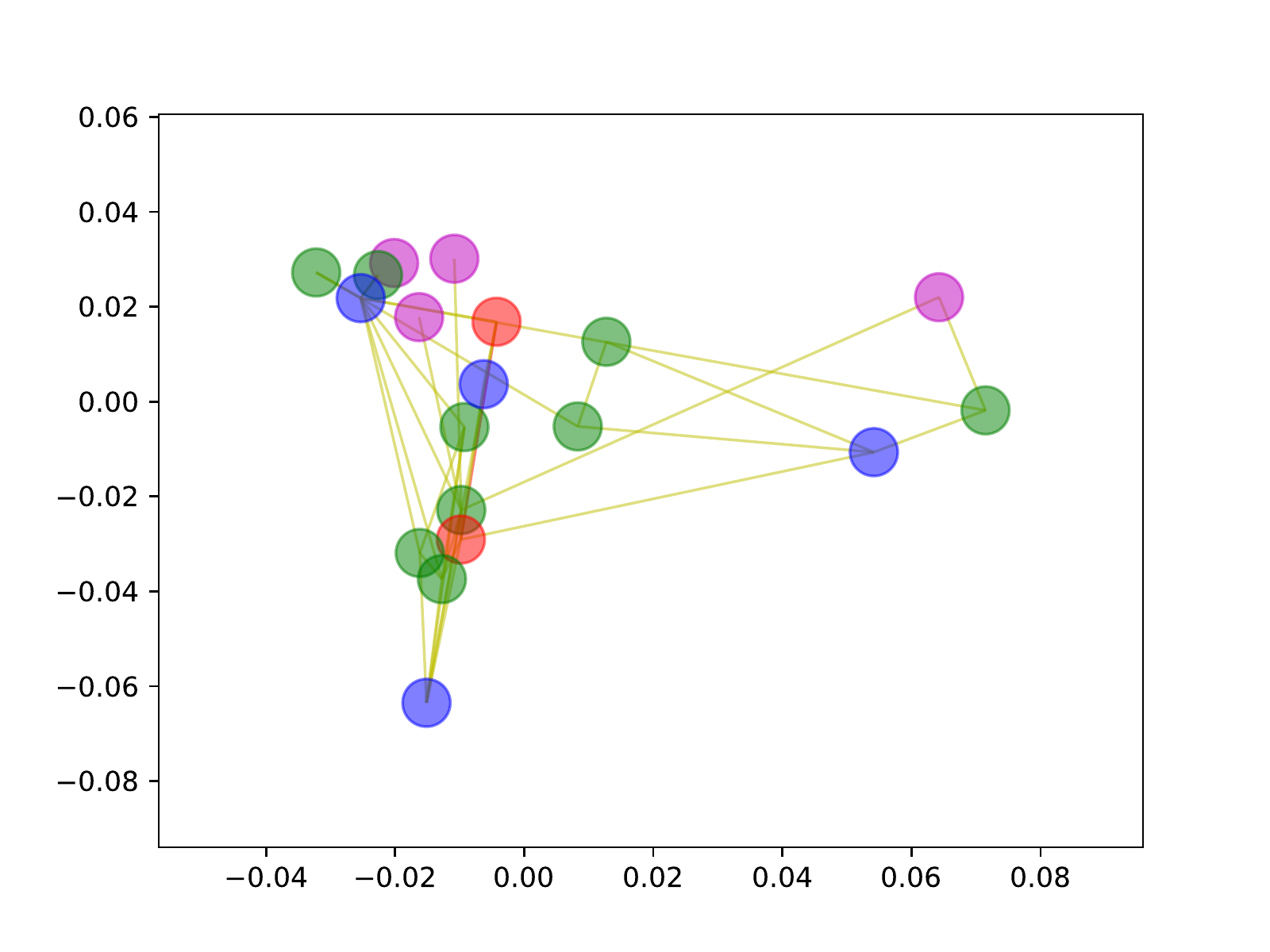}\includegraphics[width=0.25\textwidth]{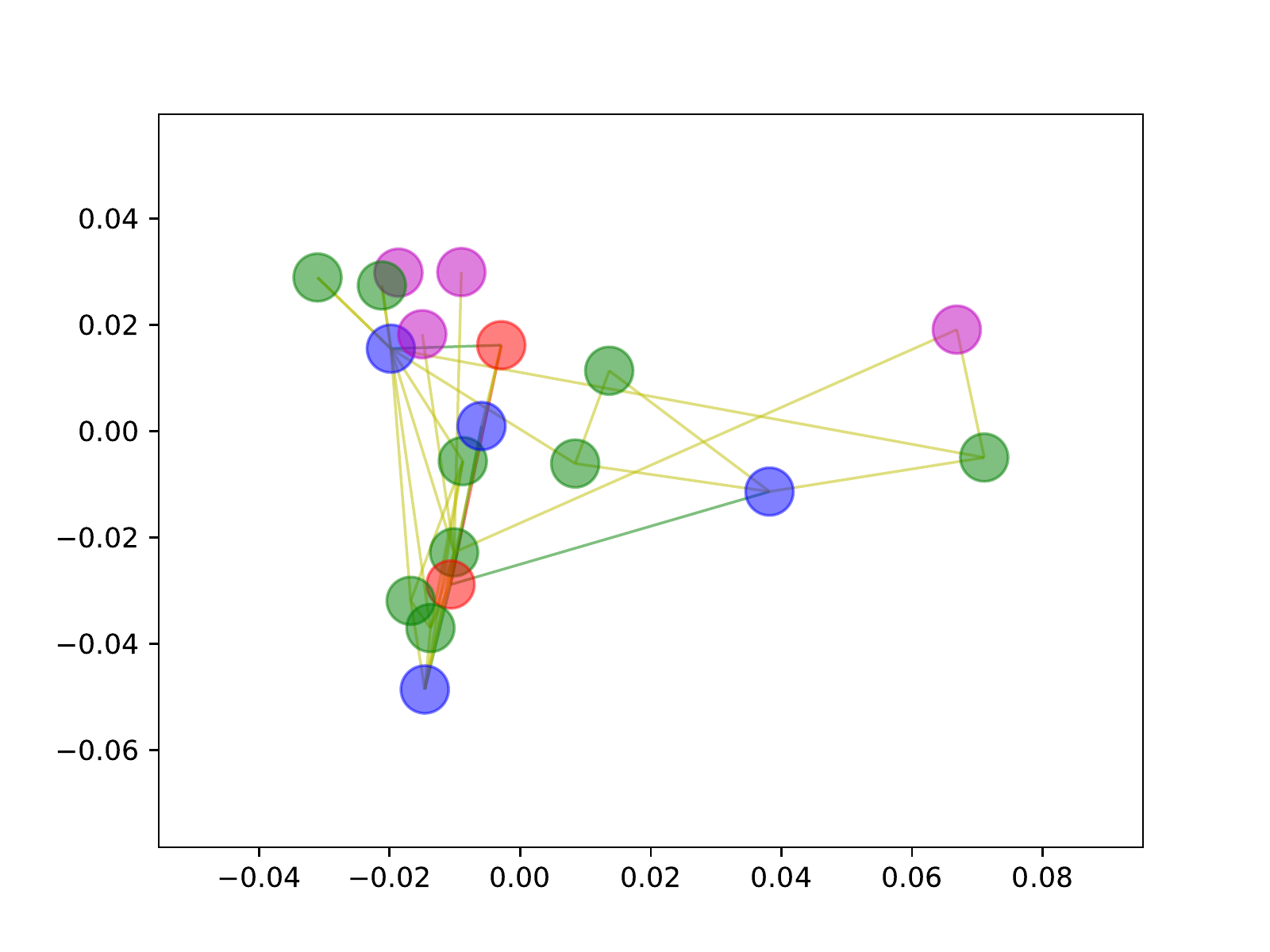}\includegraphics[width=0.25\textwidth]{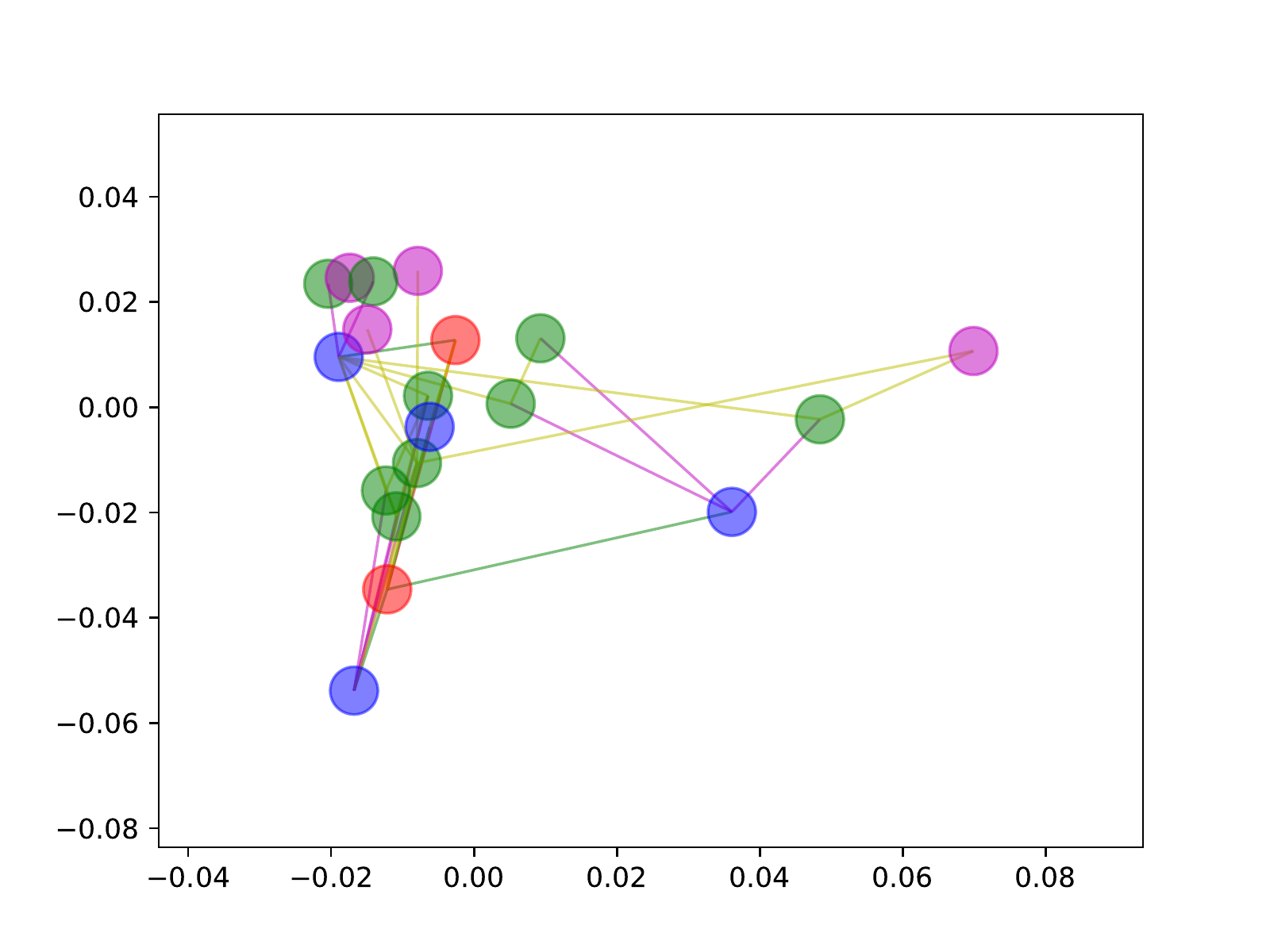}}
\end{minipage}%
}%

\subfigure[Facebook]{
\begin{minipage}[b]{1\textwidth}
\label{fig:facebook_vishops} 
\centerline{\includegraphics[width=0.25\textwidth]{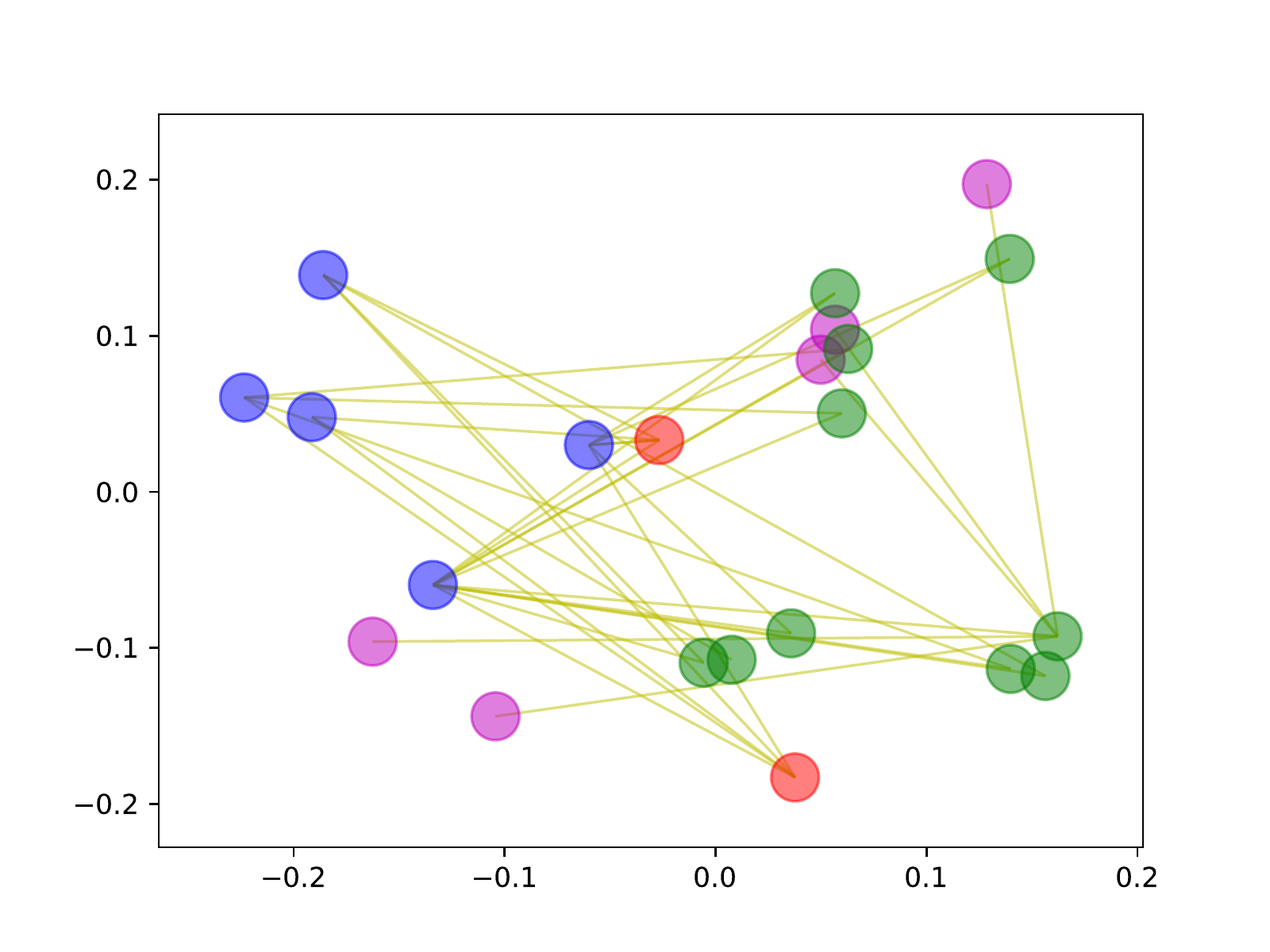}\includegraphics[width=0.25\textwidth]{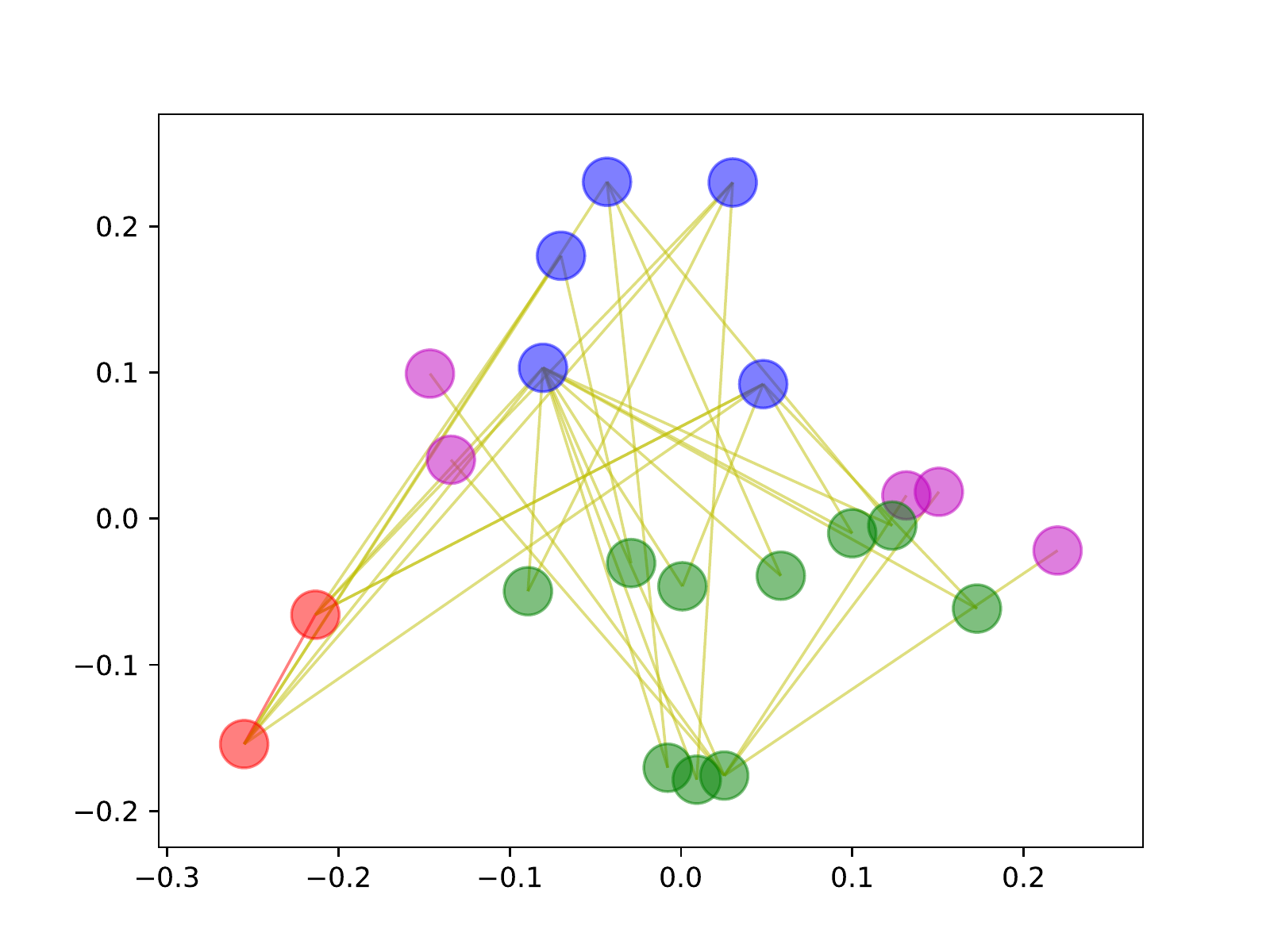}\includegraphics[width=0.25\textwidth]{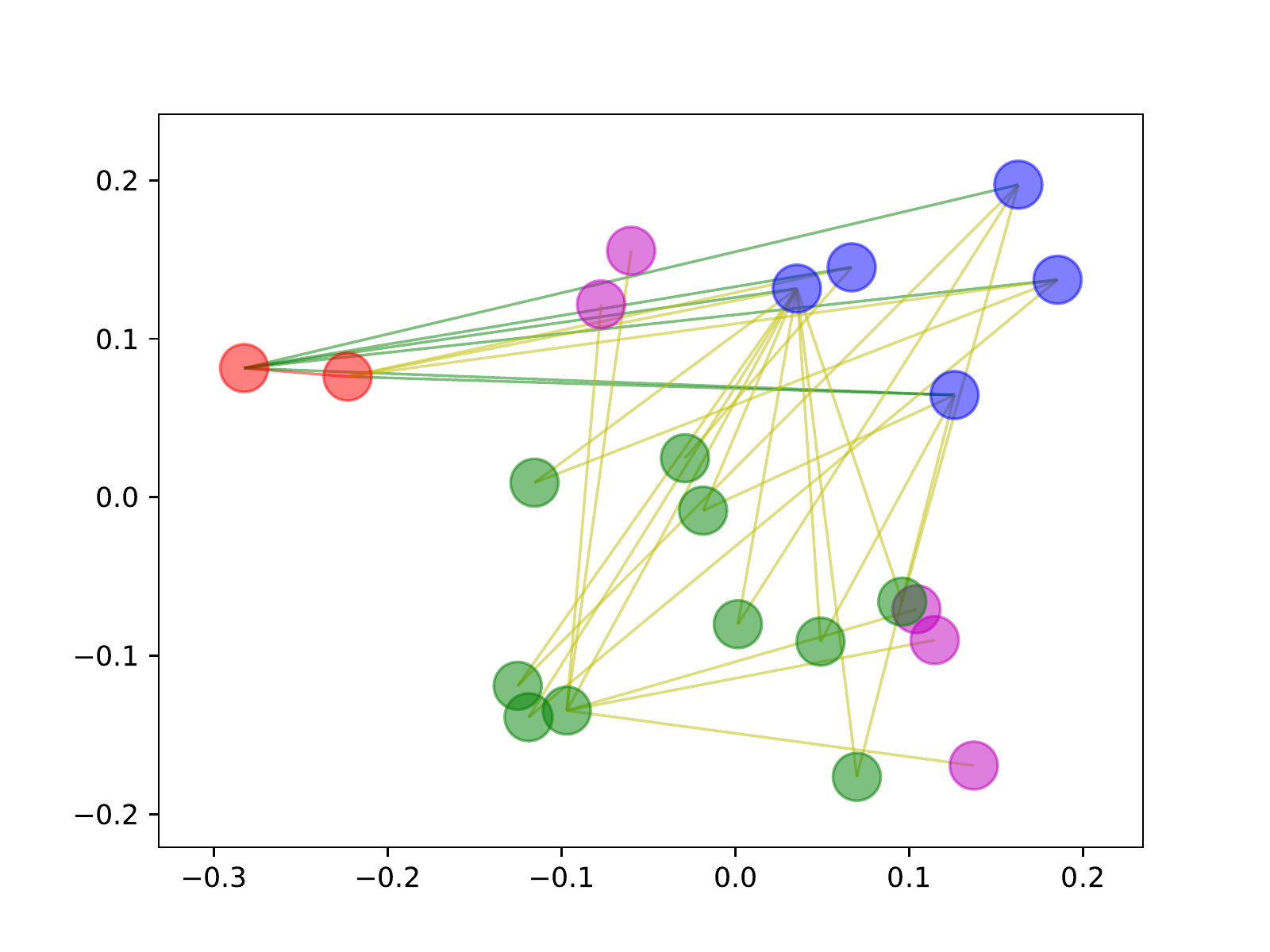}\includegraphics[width=0.25\textwidth]{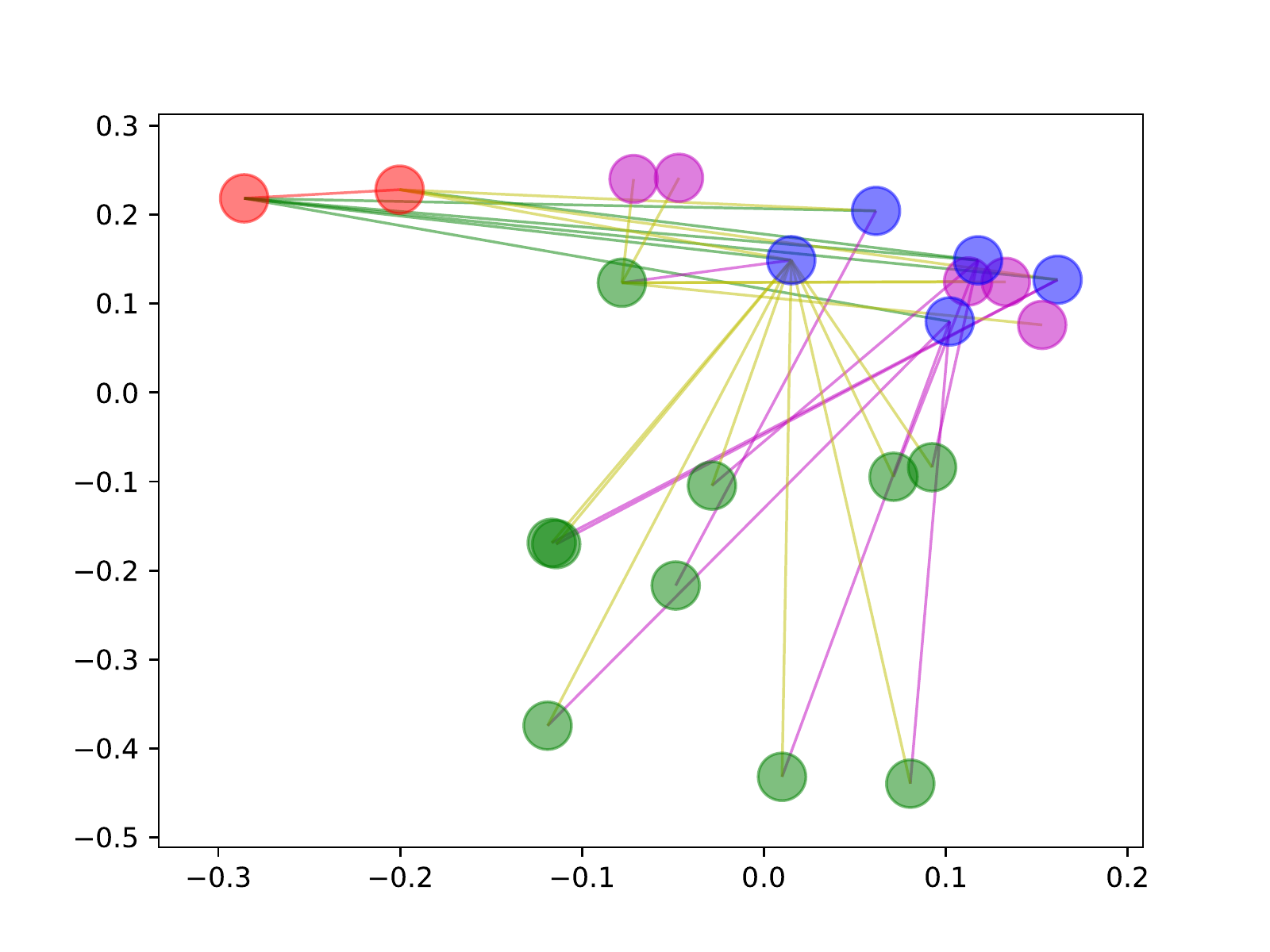}}
\end{minipage}%
}%
\caption{Visualization of high order embedding updating, from left to right are original embedding, 1st-, 2nd-, 3rd-order, correspondingly. The red nodes are nodes of the changing edge. Blue nodes are the neighbors of red nodes, and green nodes are 2-hop neighbors of red nodes. Purple nodes are more than 3 hops away from red nodes.}
\label{fig:vishops}
\end{figure*}

\begin{table}
\centering\caption{The average number of changing node in different orders of DyGCN.}
\setlength{\tabcolsep}{2.4mm}{\begin{tabular}{l|cccc|c}  
\toprule
   & 1st & 2nd & 3rd & 4th & \# total nodes\\
 \midrule

AS &  63.53 & 39.85 & 14.14 & 4.22 & 7716 \\
HEP-TH & 6.40 & 1186.58 & 1955.85 & 1658.61 & 14351\\
Facebook & 1637.69 & 546.46 & 14.14 & 0.46 & 2235\\
\bottomrule
\end{tabular}}
\label{tab:average_num_between_hops}
\end{table}

\begin{figure}[t]
\centering
\includegraphics[width=\linewidth]{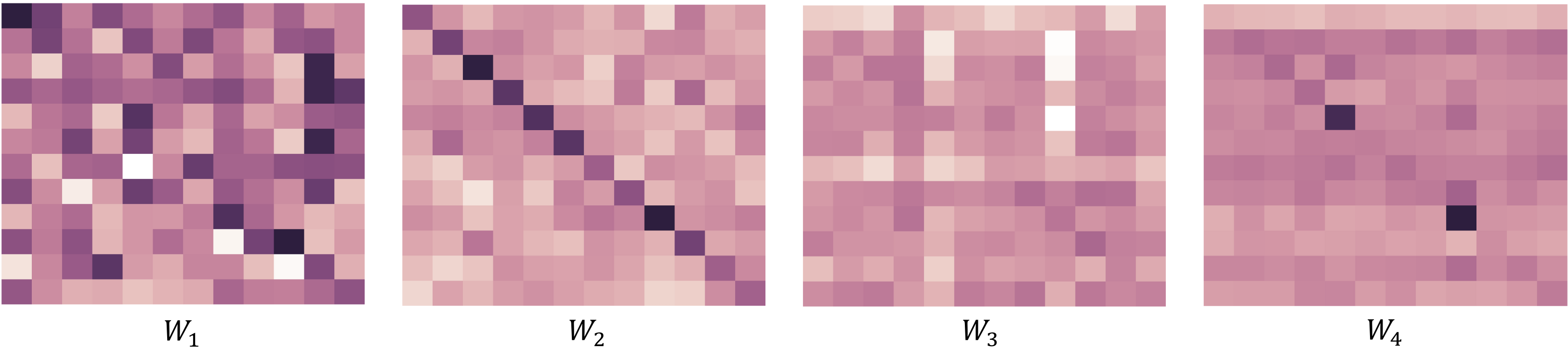}
\caption{Visualization of graph kernel matrix $W_{k}$ of DyGCN on AS dataset.}
\label{fig:gcn_w}
\end{figure}

\subsection{High Order Update Visualization}
Table \ref{tab:average_num_between_hops} summarizes the  average number of changing node in different orders of DyGCN. In this section, we visualize the update processing of node embeddings during different orders, to uncover the influence between different orders. Every node of the graph is located by 2-D vectors which is reduced by PCA.  By using the trained 4th-order DyGCN model, we randomly choose one edge as an adding edge between two snapshots. In that case, the 1st-order changing nodes are two red nodes, which is directed linked to each other. The result is shown in Figure \ref{fig:vishops}. From the first figure to the second, only the 1st-order nodes (orange) embeddings change. Among the three datasets, all the 1st-order nodes become closer with other. During the second update (blue), nodes in As and Facebook tend to be away from the  1st-order nodes, while they comes closer in HEP-TH. When the graph is dense, the information of an adding link seems only make the relationships between 2nd-order nodes close, while they may not draw closer relationships between 1st-order nodes and 2nd-order nodes. When the graph is larger, a new link may propagate the information of closer relations between 1st- and 2nd-order nodes. From the third figure to the fourth, 3rd-order nodes (green) tend to be away from the linked edges. It may further describe the reason why 2nd-order DyGCN tends to have the best performance among these three datasets.

\subsection{Graph Convolution Kernel Visualization}
To further understand what the graph convolution operators do, we visualize the weight of graph convolution kernels $W_{k}$. We take the AS dataset as an example. As is shown in Figure \ref{fig:gcn_w}, $W_{1}$ is the transform matrix between the nodes who have an additional link in at this time. While $W_{2}$, $W_{3}$, $W_{4}$ are the transform matrices of high-order update. In the figure, $W_{1}$ and $W_{2}$ have the obvious activated parts, but $W_{3}$ and $W_{4}$ seem to have low and average weights among all parts, which means $W_{1}$ and $W_{2}$ gain an exact pattern during optimization, while $W_{3}$ and $W_{4}$  seems to have difficulty in learning a distinct node propagation pattern. It can be concluded that the 1st-order and 2nd-order update have a direct and obvious patterns to learn, while model may be more difficult to learn useful patterns in 3rd-order and 4th-order update. That could be another explanation of why 2nd-order DyGCN gets the best performance.
Especially, $W_{2}$ gets a high weight in diagonal line, which means nodes tend to transform its original information to neighborhoods during their 2nd-order propagation. This insight confirms to the propagation mechanism that GCN model is supposed to have i.e., propagating its own information to the neighborhoods.

%% file: 5_conclusion.tex
In this work, we focus on how to efficiently extend GCN models to learn node embeddings in dynamic graph setting.
The proposed DyGCN naturally generalizes the embedding propagation scheme of GCN to dynamic setting in an efficient manner, which is to propagate the change along the graph to update node embeddings.
The most affected nodes are first updated, and then their change are propagated to the further nodes and leads to their update. We propose two different ways of update, i.e., high-order update and spectral propagation.
Experimental results demonstrate the efficiency and effectiveness of DyGCN and Spectral DyGCN compared with other  methods and the static counterpart (GCN).